\documentclass[final,3p,times]{elsarticle}
\usepackage{mathtools}
\usepackage{amssymb}
\usepackage{stmaryrd}
\usepackage{color}
\usepackage{pdflscape}
\usepackage{epsfig}

\usepackage{epstopdf}
\usepackage{multirow}
\usepackage{hhline}
\usepackage{txfonts}
\usepackage{mathrsfs}
\usepackage{color}
\usepackage{algorithmic}

\usepackage{mathrsfs}
\usepackage{sidecap}
\usepackage{graphicx, array}
\usepackage{makecell}
\usepackage{mathabx}

\usepackage{booktabs}

\usepackage{lineno,hyperref}
\modulolinenumbers[5]

\numberwithin{equation}{section}
\usepackage{amsmath} 
\usepackage{mathrsfs} 
\usepackage[mathscr]{euscript}
\usepackage{graphicx} 
\usepackage{subcaption}

\allowdisplaybreaks

\journal{Journal of }

\begin{document}

\begin{frontmatter}

\title{
DeepVIVONet: Using deep neural operators to optimize sensor locations with application to vortex-induced vibrations
}

\author[Brownaddress1]{Ruyin Wan}

\author[Brownaddress2]{Ehsan Kharazmi}

\author[MITaddress]{Michael S Triantafyllou}

\author[Brownaddress2]{George Em Karniadakis}

\address[Brownaddress2]{Division of Applied Mathematics, Brown University, Providence, RI 02906}
\address[Brownaddress1]{School of Engineering, Brown University, Providence, RI 02906}
\address[MITaddress]{Department of Mechanical Engineering, MIT, Cambridge, MA 02139}

\begin{abstract}
We introduce DeepVIVONet, a new framework for optimal dynamic reconstruction and forecasting of the vortex-induced vibrations (VIV) of a marine riser, using field data. We demonstrate the effectiveness of DeepVIVONet in accurately reconstructing the motion of an off--shore marine riser by using sparse spatio-temporal measurements. We also show the generalization of our model in extrapolating to other flow conditions via transfer learning, underscoring its potential to streamline operational efficiency and enhance predictive accuracy. The trained DeepVIVONet serves as a fast and accurate surrogate model for the marine riser, which we use in an outer--loop optimization algorithm to obtain the optimal locations for placing the sensors. 
Furthermore, we employ an existing sensor placement method based on proper orthogonal decomposition (POD) to compare with our data-driven approach. We find that that while POD offers a good approach for initial sensor placement, DeepVIVONet's adaptive capabilities yield more precise and cost-effective configurations.

\end{abstract}

\begin{keyword}
Deep Operator Network; Vortex induced vibrations; Transfer learning; Sensor placement
\end{keyword}

\end{frontmatter}

\thispagestyle{plain}

\section{Introduction}
The phenomenon of vortex-induced vibrations (VIV) presents significant challenges in offshore engineering, particularly concerning the dynamic motions of marine risers and the formation of large-scale vortices in their wake. Marine risers are critical components used to transport fluids between seabed wells and floating platforms. Understanding and predicting VIV is essential due to its implications for the structural integrity and operational safety of offshore platforms. In recent years, Computational Fluid Dynamics (CFD) methods such as DNS \cite{bourguet2011vortex}, LES \cite{wang2021large}, RANS \cite{zheng2022flow}, SPH \cite{sun2021accurate}, and DVM \cite{lin2019numerical, jin2023numerical} have been extensively employed to investigate VIV in flexible cylinders. These simulations have provided significant insights into the wake dynamics and the complex fluid-structure interaction mechanisms but at low speeds. However, the inherent complexity of VIV interactions and high Reynolds number complicates accurate predictions, making traditional CFD infeasible, while experimental methodologies entail significant logistical challenges and high costs \cite{williamson1996vortex,sarpkaya2006viv}. Several papers have reviewed these challenges in VIV \cite{bearman1984vortex,wang2020review,ma2022flexible} in understanding  interaction mechanisms and developing reliable prediction methods.

Recent advancements in machine learning have shown promise across various fields \cite{jin2020sympnets, zhang2022aoslo, toscano2023teeth, daneker2022systems, raissi2019physics, zhang2021integrated, ovadia2023realtime, zhang2022analyses, chen2020physics} and especially engineering problems in fluids, including flow field prediction and shape optimization \cite{shukla2024deep}, turbulence modeling \cite{oommen2023integrating}, and solving Riemann problems in compressible flows for extreme pressure jumps \cite{peyvan2024riemannonets}. More developments have also been made in the field of scientific machine learning with focus on solving PDEs and domain decomposition \cite{kharazmi2019variational, kharazmi2021hp,jagtap2020conservative}. For offshore structures in particular, Cao et al. \cite{cao2023deep} have utilized different neural operators in capturing the responses of a floating structure. To address the complexities associated with VIV, Kharazmi et al.\cite{kharazmi2021inferring, kharazmi2021data} have used physics-informed neural networks (PINNs) to infer the the motion of VIV given data on displacement and forces, Mentzelopoulos et al. \cite{mentzelopoulos2024variational} have used transformer to forecast the the vibrations in time using sparse vibrations. Notably, the Deep Neural Operator (DeepONet), which utilizes two sub-networks, a branch net for encoding input functions and a trunk net for encoding domain geometry, has been effective in capturing the dynamics of complex systems from sparse measurements (\cite{lu2021deeponet}). Deep neural operators are particularly suited for problems where traditional neural network architectures struggle to generalize across different functional inputs.

In this context, we introduce DeepVIVONet, a novel approach based on the DeepONet architecture designed specifically for the data-driven inference of VIV. We present the formulation of DeepVIVONet to leverage sparse spatial and dense temporal measurements for dynamic prediction of VIV. We begin by detailing the problem setup and the unique architecture of DeepVIVONet, emphasizing its capability to integrate sparse observational data effectively.
Following the architectural overview, we demonstrate the application of DeepVIVONet in predicting the dynamics of marine risers across both current and future states using a limited number of observer sensors. This approach not only enhances the understanding of VIV phenomena but also optimizes the placement and use of sensors, potentially reducing the overall costs associated with traditional experimental methodologies.
Moreover, we explore the scalability and adaptability of DeepVIVONet through transfer learning techniques, assessing its ability to extrapolate and apply learned dynamics to different but related scenarios. This aspect of the research underscores the flexibility and robustness of DeepVIVONet in handling various operational environments and conditions.
Finally, we introduce a methodological framework for the strategic placement of observer sensors, which is critical for maximizing data acquisition efficiency and model performance. This framework is designed to support offshore engineers in making informed decisions regarding sensor deployment, thereby enhancing the overall predictive capability and reliability of marine riser operations.

In summary, this paper contributes to the ongoing efforts to refine predictive modeling techniques for VIV, with the potential to impact the field of offshore engineering by providing a more quantitative  understanding of complex fluid-structure interactions and improving the operational safety and efficiency of marine infrastructure.

\label{Sec: introduction}

%
\section{DeepVIVONet: Deep Operator Network for VIV}


Here, we develop a deep neural operator approach for predicting the vortex-induced vibration (VIV). From the theoretical standpoint, it was proved that the neural networks (NNs) not only can be used to approximate any continuous function \cite{Hornik1989}, but also approximate accurately any continuous nonlinear functional \cite{chen1995universal} or operator (a mapping from a function to another function) \cite{chen1995universal}. Based on the approximation theorem for operators, a specific network architecture, namely the deep operator network (DeepONet) was proposed in \cite{lu2021deeponet}, which also presented a theoretical analysis and demonstrated various examples to show the accuracy and high efficiency of this formulation. Specifically, it was shown that DeepONet can learn both explicit as well as implicit operators, e.g., in the form of PDEs. 

Next, we first give a brief overview of DeepONet as the building block for the current development, and then we discuss the problem set up,  specifically for the VIV problem.

\subsection{DeepONet: a brief overview}

\begin{figure}[h!]
\centering
\includegraphics[width=0.6\linewidth]{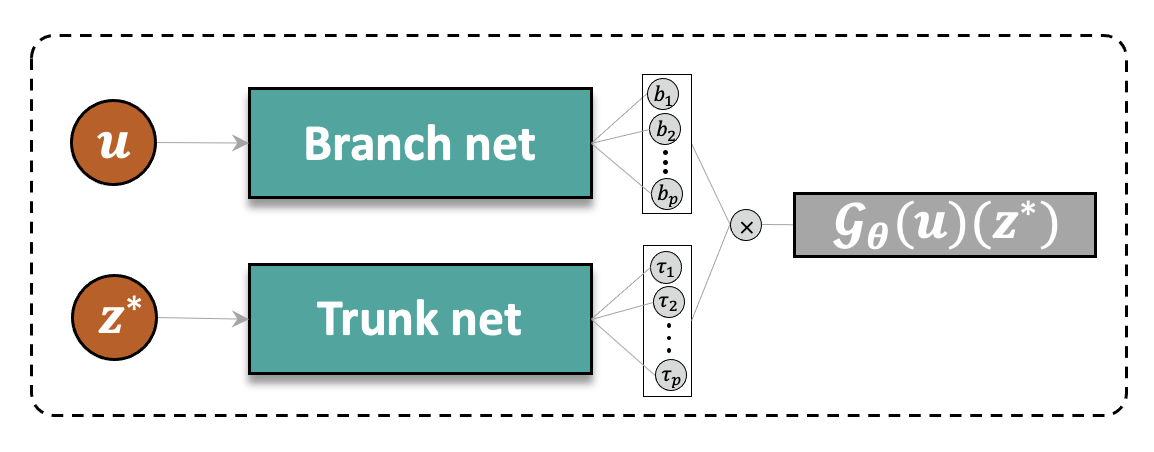}
\caption{DeepONet architecture. This unique architecture is comprised of two primary networks, namely, Trunk Network (TN) and Branch Network (BN). It forms a mapping from the input $u$ to the BN to the output evaluated at input $z^\star$ to the TN.}
\label{fig:DeepONet}
\end{figure}

DeepONet~\cite{lu2021deeponet} represents a cutting-edge paradigm in machine learning designed specifically for learning mappings from functions to functions. This architecture comprises two primary subnetworks: the branch net (BN) and the trunk net (TN); see Fig.\ref{fig:DeepONet}. 
The branch net processes inputs $ u $ evaluated at discrete points set $ z = \lbrace z_{1}, z_{2}, \cdots, z_{m} \rbrace$within an infinite-dimensional functional space $ U $, while the trunk net operates on the coordinates $ \lbrace z^{\star}_1, z^{\star}_2, \cdots, z^{\star}_p \rbrace$, where the output function is to be evaluated. Together, these components enable DeepONets to model complex nonlinear operators efficiently. DeepONet is based on the universal operator approximation theorem \cite{chen1995universal}, ensuring its capability to approximate arbitrary functions through a learned mapping $ \mathcal{G} $. The output function $ \mathcal{G_{\theta}}(u)(z^{\star}) $ is expressed as:
\begin{align}
\label{eq: deeponet formula}
\mathcal{G_{\theta}}(u)(z^{\star}) = \sum_{k=1}^{P} B_k(u) T_k(z^{\star}) + B_0,
\end{align}
where $ B_k(u) $ and $ T_k(z^{\star}) $ represent outputs from the branch and trunk nets, respectively, while $ B_0 $ serves as a bias term.

To train a DeepONet, we use labeled data consisting of inputs $\lbrace u^{(i)} (z), z^{\star} \rbrace_{i = 1, \cdots, N_{obs}}$ and outputs $\lbrace o^{(i)} (z^{\star}) \rbrace_{i = 1, \cdots, N_{obs}}$. The output from the trained DeepONet, $\lbrace \mathcal{G_{\theta}}(u)^{(i)} (z^{\star}) \rbrace_{i = 1, \cdots, N_{obs}}$ is the approximation of the output labels. In such a case, the network parameters $\theta$ are obtained by minimizing the empirical loss function defined based on the mean square error between the true $\lbrace \mathcal{G}(u)^{(i)} (z^{\star}) \rbrace_{i = 1, \cdots, N_{obs}}$ and the network prediction $\lbrace \mathcal{G_{\theta}}(u)^{(i)} (z^{\star}) \rbrace_{i = 1, \cdots, N_{obs}}$: 
\[
\mathcal{L}(\theta) = \frac{1}{N_{obs}} \sum_{i=1}^{N_{obs}} (\mathcal{G_{\theta}}(u)^{(i)} (z^{\star}) - \mathcal{G}(u)^{(i)} (z^{\star}))^2 = \frac{1}{N_{obs}} \sum_{i=1}^{N_{obs}} (\mathcal{G_{\theta}}(u)^{(i)} (z^{\star}) -o^{(i)} (z^{\star}))^2,
\]
This efficient structure makes DeepONet ideal for dynamic system reconstruction where input-output mappings may vary significantly.

\subsection{DeepONet: architecture and data structure specifically for dynamic reconstruction and forecasting}
We assume that we have a system whose dynamics is described by $u(t,x)$ over a time-space domain $\Omega$, and follows an underlying physical law  described by a system of differential equations, e.g., as $$ \mathcal{G}(u) = F(t,u),  $$ where the operator $\mathcal{G}$ is comprised of temporal and spatial derivatives such as $ \frac{\partial u}{\partial t}$ and $\nabla u$. We assume that we have a few measurements of the dynamics $u$ at some (usually sparse) points inside the domain $\Omega$. The objective in the current setup of the problem is to build a surrogate of the system that is capable of \emph{reconstructing} a continuous representation of the dynamics inside the domain $\Omega$ and \emph{predicting/forecasting} outside of the domain. 

In particular, let $\Tilde{\mathcal{G}}$ be the surrogate of the system, and $\Tilde{u}$ be the point measurement of the system states on the points $\Tilde{z} \in \Tilde{\Omega} \subset \Omega$. Thus, the surrogate  $\Tilde{\mathcal{G}}$ is the mapping of the dynamic discrete points measurement of $\Tilde{u}$ from the subdomain $\Tilde{\Omega}$ to the continuous output over the entire domain $\Omega$, $$\Tilde{\mathcal{G}}:  \tilde{u}|_{\Tilde{z}} \,\, \rightarrow \,\, u|_{z^\star}. $$ The DeepONet architecture explained above serves as the surrogate $\Tilde{\mathcal{G}}$ to parametrize this mapping via a unique architecture of neural networks. Therefore, we have: $\mathcal{G} \approx \Tilde{\mathcal{G}} = \mathcal{G}_{\theta}$, where $\mathcal{G}_{\theta}$ is the DeepONet in equation \ref{eq: deeponet formula}.

\subsection{DeepVIVONet: DeepONet setup specific to the VIV problem}

A marine riser is submerged into deep ocean and experiences complex dynamic responses due to ocean currents and fluid-structure interactions. In this context, we consider the dynamics of the riser as a continuous system expressed by a variable as a function of time and space, governed by an underlying physical model. The dynamics leads to in-line (IL) and cross-flow (CF) strains, denoted by $\varepsilon_x(z,t)$ and $\varepsilon_y(z,t)$, respectively, where the $z$ coordinate is along the structure and the $x$ and $y$ coordinates are in the IL and CF directions, respectively. The dynamics of the riser is described by a linear beam-string equation as follows, 
\begin{align}
\label{eqn: BSE}
\frac{\partial^2 \varepsilon}{\partial t^2}
+\zeta\frac{\partial \varepsilon}{\partial t}
+\mathrm{EI} \frac{\partial ^4 \varepsilon}{\partial z^4}
-\mathrm{T} \frac{\partial ^2 \varepsilon}{\partial z^2}=F,
\end{align}
where $\varepsilon$ is either the CF or IL displacements and $F$ is the lift and drag, respectively. The parameter $\zeta$ is the damping ratio, $\mathrm{T}$ is the tension and $\mathrm{EI}$ is the bending stiffness; all normalized by the riser mass per unit length.

\begin{figure}[h!]
\centering
\includegraphics[clip, trim = 3cm 1cm 2.5cm 1cm, width=0.6\linewidth]{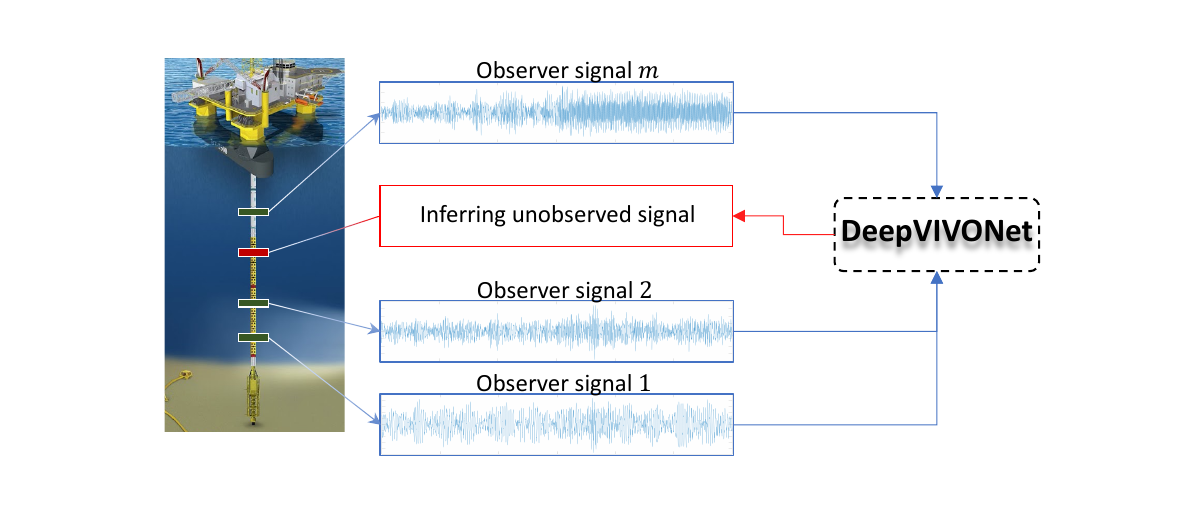}
\caption{DeepVIVONet in VIV problem: DeepVIVONet is based on DeepONet for the VIV problem. This framework reconstructs the dynamics of a marine riser by using only a few measurements from ``observer" sensors. It essentially builds a mapping from observer data to the entire domain. }
\label{fig:DeepVIVONet}
\end{figure}

The objective is to infer the full dynamic behavior of the riser using sparse spatial measurements that are relatively dense in time. Specifically, we assume that there are $m$ ``observer" sensors instrumented along the riser structure. These observer sensors continuously stream the measurements in time with a fixed measurement frequency $\mathcal{F}$. In this setup, we train our surrogate, i.e., DeepVIVONet, to reconstruct the entire dynamics in space--time domain using the observer signals as input. The schematic of DeepVIVONet is shown in Fig. \ref{fig:DeepVIVONet}.  
The blue color signals are the streaming signals from observer sensors that are fed into the DeepVIVONet as input. The DeepVIVONet then outputs the red color signal at any arbitrary location along the riser. This is a unique architecture leveraged by using DeepONet, since the output of DeepVIVONet is continuous function in space and thus it can predict the signal at any $z$ location. Thus, during training, we use the label data to train our model. After training and during inference, the trained model predicts the signal at any location and therefore provides a continuous reconstruction of the dynamic in space. 

In the proposed architecture of DeepVIVONet, our model takes the history of observer signals as input and predict the dynamics at the current time step and also future time step. It essentially provides a forecasting of the reconstructed dynamics in time, forming a \textit{reconstruction and forecasting} scheme. To achieve this, we follow a specific pattern to structure our training data that is consistent with the architecture of our DeepVIVONet and our objectives.

\subsection{DeepVIVONet: architecture and data structure}
\label{subsec: DeepVIVONet architecture}

\begin{figure}[h!]
\centering
\includegraphics[clip, trim = 0cm 0cm 0cm 0cm, width=1\linewidth]{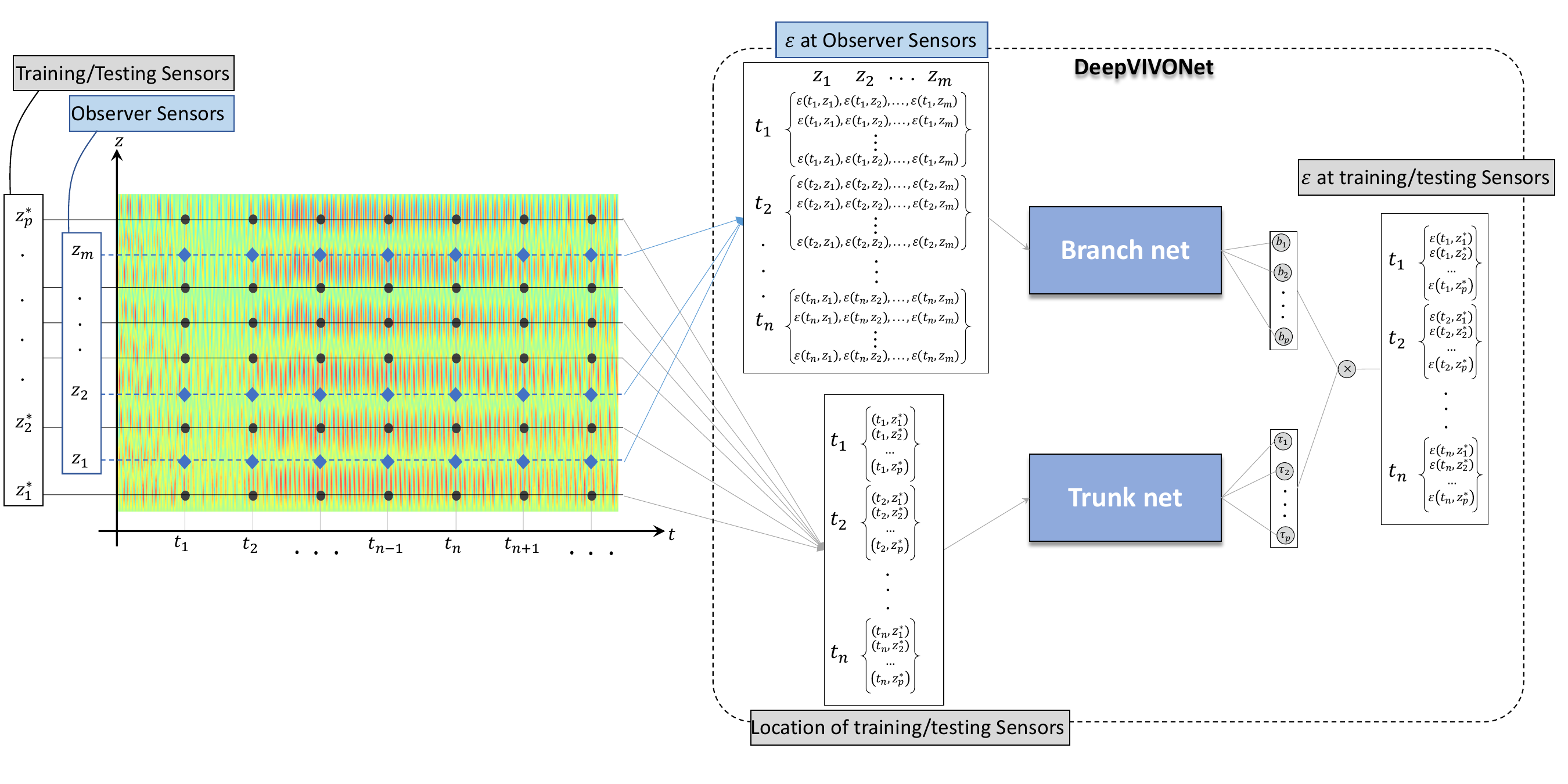}
\caption{Detailed structure of input and output data for the DeepVIVONet framework. Left: the background shows the actual space--time data. We assume that we have $m$ observers shown with blue diamonds and we have $p$ sensors for training/testing data.}
\label{fig:DeepVIVONet-data}
\end{figure}

DeepVIVONet is comprised of two fully connected networks as branch and trunk nets, each with separate number of layers, neurons, and activation functions. The detailed architecture of DeepVIVONet is shown schematically in Fig. \ref{fig:DeepVIVONet-data} (right side). The input to the branch net is the history of observer signals $\varepsilon$ instrumented at locations $z_1, \cdots, z_m$. The input to the trunk net is the location at which the output signals are computed, i.e., the location of labeled data during the training and the location of point of interest during inference.

In Fig. \ref{fig:DeepVIVONet-data} (left side), we show a schematic of data to explain the structure of input/output for DeepVIVONet. We assume that the entire data is given at $m+p$ points sparse in space but dense discrete measurements in time. We divide the data in time in two distinct time windows: the training window and the forecasting window. In both time windows, we use the $m$ observers as input, however, in the training window we use the other $p$ sensors as labeled data to train our model, and in the forecasting window we use them to test our model. We see from Fig. \ref{fig:DeepVIVONet-data} that at any particular time step $t_j$, there are $m$ observers located at $\lbrace z_1, z_2, \cdots, z_m \rbrace$ (marked by blue diamonds). These points are the input to the branch net at the time step $t_j$. The $z^*$ values of the other $p$ sensors at $\lbrace z^{*}_1, z^{*}_2, \cdots, z^{*}_p \rbrace$ (marked by black circles) are the input to the trunk net. Thus, the output of DeepVIVONet are the prediction of the signals at $z^*$ in the current and future time step, where we have the labeled data from the training sensors. By stacking the history of observer signals at observer location and future of training signals at training location, we build our training/testing datasets. 

In the \textit{reconstruction stage}, the network output is formulated as 
$$\mathcal{G_{\theta}}(\varepsilon)(t_j, z^*) = \sum_{k=1}^{P} {B_k}(\lbrace \varepsilon(t_j, z_1), \cdots, \varepsilon(t_j, z_m) \rbrace){T_k}(t_j, z^{*}) + {B_0},$$
where the measurements $\varepsilon(t_j, z^*)$ from the additional training sensors are used as labeled data to compute the loss function $L = |\mathcal{G_{\theta}}(\varepsilon)(t_j, z^*) - \varepsilon(t_j, z^*) |^2$ to train DeepVIVONet.

In the \textit{forecasting stage}, we use the stacked data to march in time. Thus, we feed the network with a stack of data from several 
look-back time steps $lb$ up to time step $j-1$. In this setup, $B_k$ with its new input is 

$${B_k}(\lbrace \lbrace \varepsilon(t_{j-lb}, z_1), \cdots, \varepsilon(t_{j-lb}, z_m) \rbrace, \cdots, \lbrace \varepsilon(t_{j-1}, z_1)\rbrace \rbrace)$$

and thus, the network learns the forecasting mapping from a look-back window to the future time steps.

\section{Results on Dynamic Reconstruction and Forecasting}

%

\begin{figure}[h!]
\centering
\includegraphics[clip, trim = 0cm 0cm 0cm 0cm, width=0.6\linewidth]{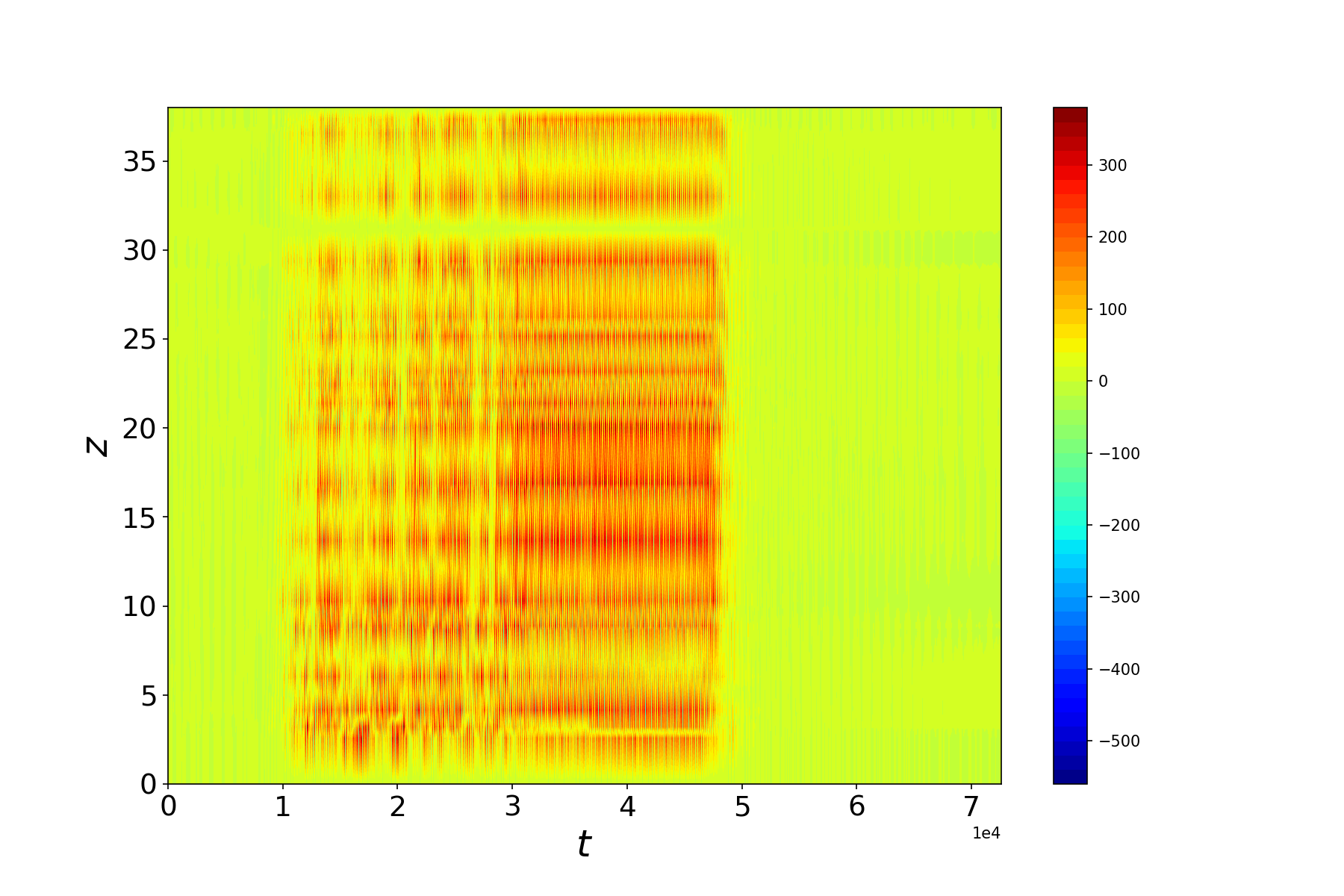}
\caption{NDP data set: CF strain for shear flow case ``test2430". The vertical axis z is the distance from the bottom of the riser in meter. The horizontal axis is the time step at which the data has been collected with the frequency of 1200 Hz. }
\label{fig: NDP shear}
\end{figure}

In this section, we present the results of DeepVIVONet in reconstruction and forecasting of the riser dynamics using data from the Norwegian Deep water Program (NDP)~\cite{henning2005ndp}. This dataset includes the experimental results for several cases of uniform and linear shear flows. The details of test cases and sensor instrumentation are provided in Table \ref{tab: NDP cases details} in \ref{App: NDP}. A series of analysis and investigations were performed on the dataset \cite{trygve2005ndpreponse,trygve2005ndp, gro2005ndp} to prioritize future testing of long riser models that advance the understanding of VIV. Here, we choose the shear flow cases named as ``test2330", ``test2430", and ``test2500" in the dataset. These cases are based on the slow, medium, and high velocity regimes in which the maximum velocity is $ U = 0.50 \, \text{m/s}$, $U = 1.50 \, \text{m/s}$, and $ U = 2.20 \, \text{m/s}$, respectively. For the CF strain signals, there are 24 sensors along the riser (we note that the sensor 22, positioned at \( z = 31.191 \, \text{m} \), has incorrect record of strain readings and was consequently excluded from our training dataset, and thus we have a total of 23 sensors). The riser has pinned boundary conditions and thus we add two imaginary sensors with zero values at the two ends of the riser. We use all the \( p = 25 \) sensors as training sensors and select \( m = 3 \) observer sensors. We also normalize the data by scaling them down by the factor of \( 0.2 \times \max(\varepsilon(z,t)) \). The data for the case ``test2430" is shown in Fig. \ref{fig: NDP shear} as an example. The figure shows the contour plot of the CF strain. The vertical axis z is the distance from the bottom of the riser (in meters). The horizontal axis is the time step at which the data has been collected with the frequency of 1200 Hz.

\subsection{Reconstruction and forecasting of strain signals}

\begin{figure}[h!]
\centering
\includegraphics[clip, trim = 0cm 0cm 0cm 0cm, width=1\linewidth]{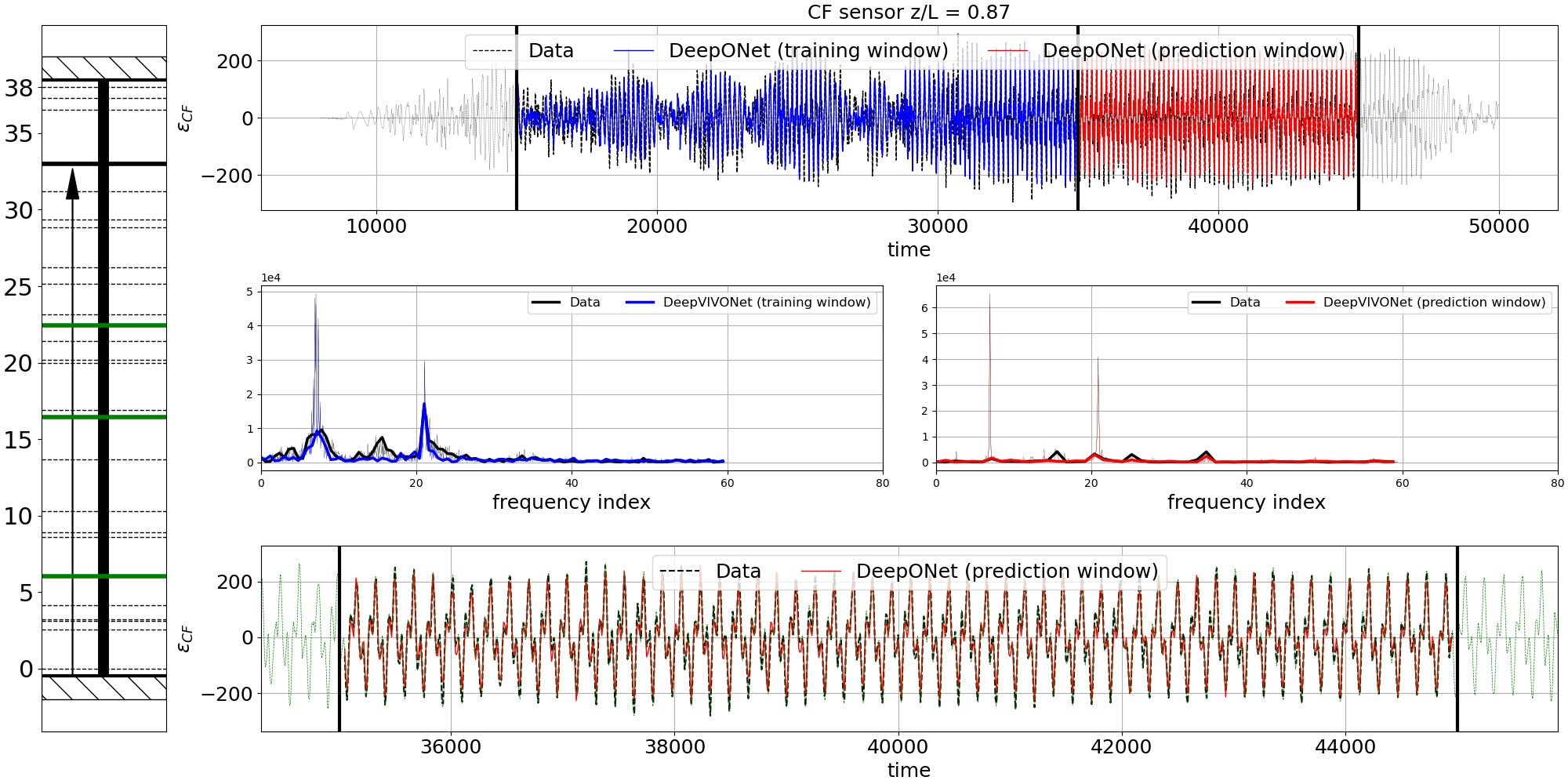}
\caption{DeepVIVONet: Prediction of CF strain for shear flow case ``test2430". Left panel: the location of all sensors (dashed line), observers (green lines), and test sensor (arrow). Top panel: data (black), DeepONet training (blue) in the training window, and  DeepONet prediction (red) in the prediction window. Middle panel: the FFT of data and DeepONet. Bottom panel: the zoomed-in plot of the prediction window.}
\label{fig: DeepVIVONet shear 2430}
\end{figure}

\begin{figure}[h!]
\centering
\includegraphics[clip, trim = 0cm 0cm 0cm 0cm, width=1\linewidth]{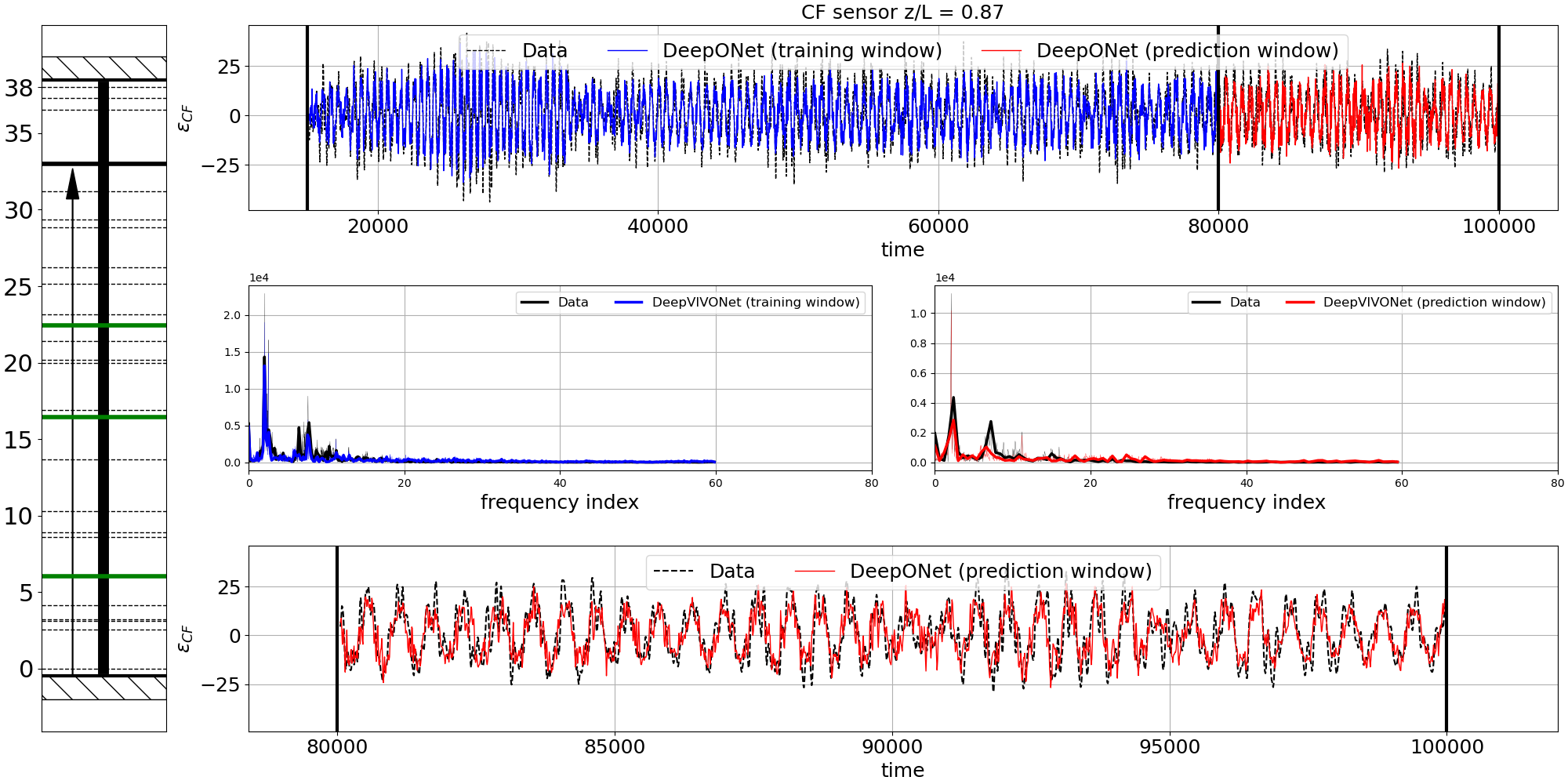}
\caption{DeepVIVONet: Prediction of CF strain for shear flow case ``test2330". Left panel: the location of all sensors (dashed line), observers (green lines), and test sensor (arrow). Top panel: data (black), DeepONet training (blue) in the training window, and  DeepONet prediction (red) in the prediction window. Middle panel: the FFT of data and DeepONet. Bottom panel: the zoomed-in plot of the prediction window.}
\label{fig: DeepVIVONet shear 2330}
\end{figure}

\begin{figure}[h!]
\centering
\includegraphics[clip, trim = 0cm 0cm 0cm 0cm, width=1\linewidth]{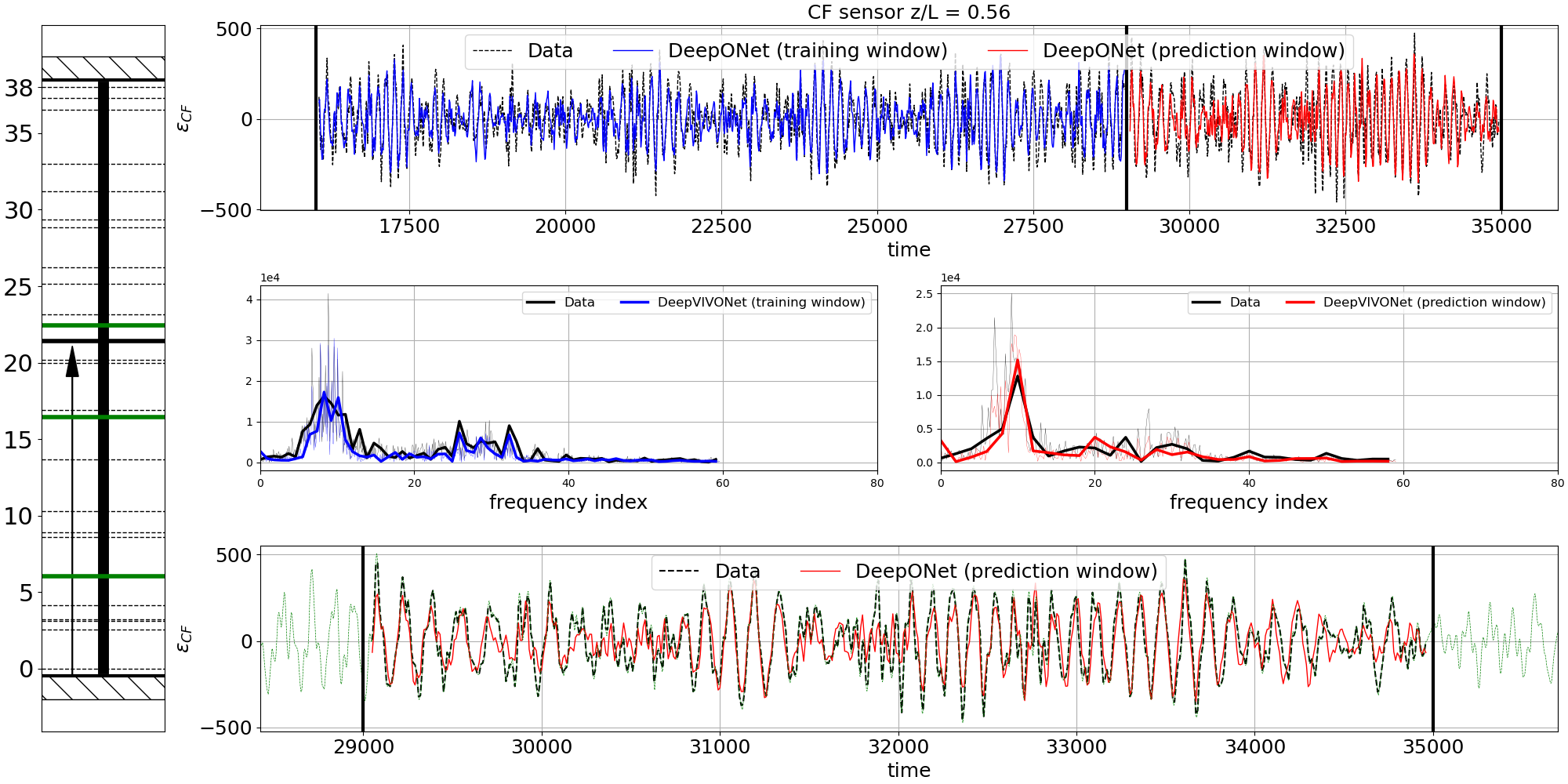}
\caption{DeepVIVONet: Prediction of CF strain for shear flow case ``test2500". Left panel: the location of all sensors (dashed line), observers (green lines), and test sensor (arrow). Top panel: data (black), DeepONet training (blue) in the training window, and  DeepONet prediction (red) in the prediction window. Middle panel: the FFT of data and DeepONet. Bottom panel: the zoomed-in plot of the prediction window.   }
\label{fig: DeepVIVONet shear 2500}
\end{figure}

Table \ref{CF shear flow prediction setup} presents the setups for CF cases across regimes with varying maximum velocities. The data was divided into training and prediction windows, and the table outlines the specific training and prediction windows for each case. The results for these cases are shown in Figures \ref{fig: DeepVIVONet shear 2430}, \ref{fig: DeepVIVONet shear 2330}, and \ref{fig: DeepVIVONet shear 2500}.

\begin{table}[h!]
\centering
\caption{Setups for different CF shear flow cases.}\label{CF shear flow prediction setup}
\begin{tabular}{cccc}
\hline
Test case  & Maximum velocity (m/s) & Training window & Test window \\ \hline
2330       & 0.50                   & [15000, 80000] & [80000, 100000] \\
2430       & 1.50                   & [15000, 35000] & [35000, 45000] \\
2500       & 2.20                   & [16000, 29000] & [29000, 35000] \\
\hline
\end{tabular}
\end{table}

\begin{figure}[h!]
\centering
\includegraphics[clip, trim = 0cm 0cm 0cm 0cm, width=1\linewidth]{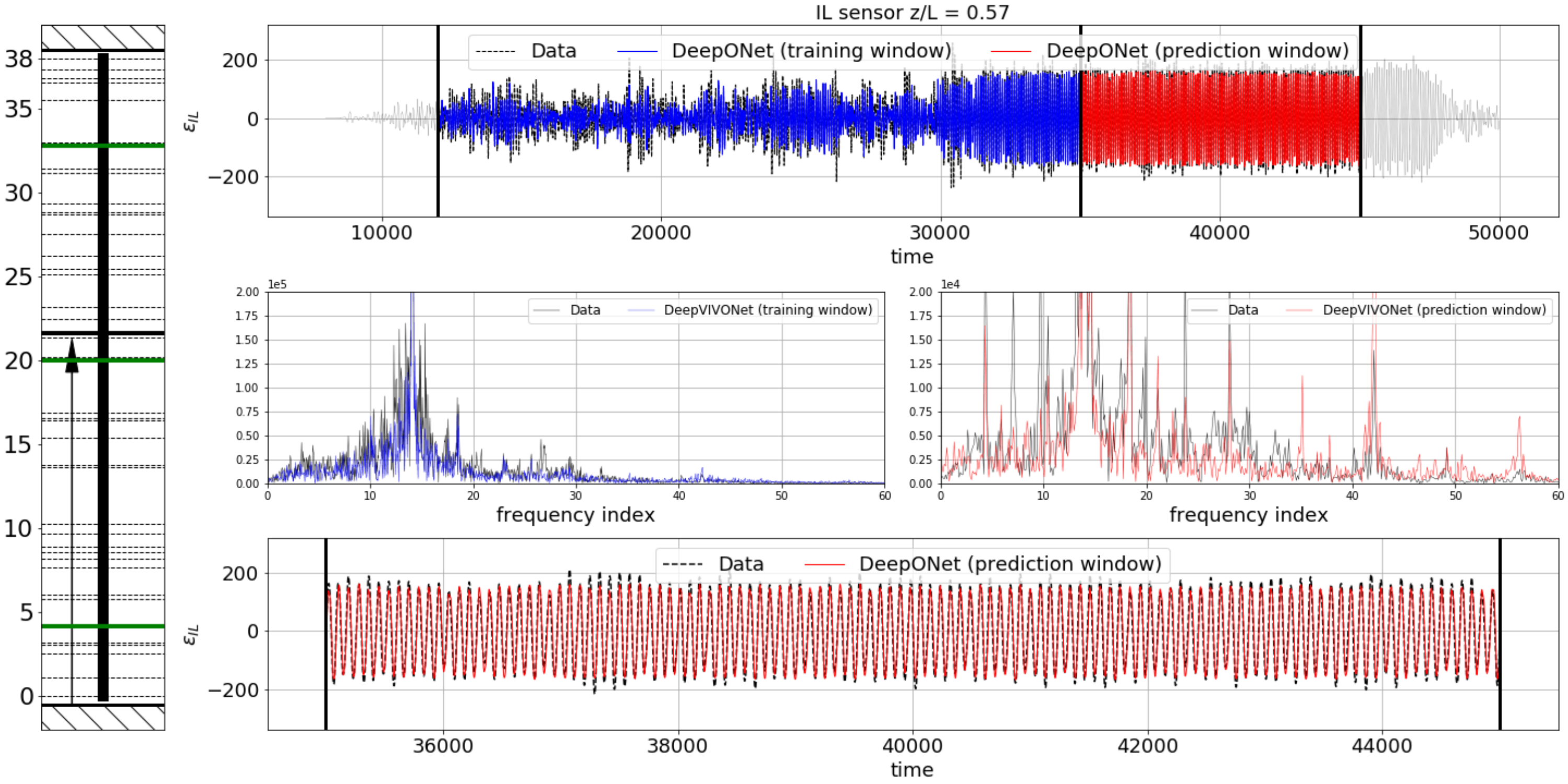}
\caption{DeepVIVONet: Prediction of IL strain for shear flow case ``test2430". Left panel: the location of all sensors (dashed line), observers (green lines), and test sensor (arrow). Top panel: data (black), DeepONet training (blue) in the training window, and  DeepONet prediction (red) in the prediction window. Middle panel: the FFT of data and DeepONet. Bottom panel: the zoom--in plot of the prediction window.}
\label{fig: DeepVIVONet shear 2430 IL}
\end{figure}

We give a detailed explanation by using the shear flow case ``test2430" as an example. For this specific test case, the data is divided into training and prediction windows, covering the time domains $[15000, 35000]$ and $[35000, 45000]$ time steps, respectively. For both the branch and trunk networks, we employ 6 hidden layers with 50 neurons in each layer. Figure \ref{fig: DeepVIVONet shear 2430} provides a comprehensive visualization of the setup and results. The left panel depicts a schematic of the riser, with horizontal dashed lines marking the locations of all instrumented sensors. Observer sensors are highlighted in green, and a black arrow points to an arbitrary location on the riser for prediction, demonstrating the model's ability to perform continuous spatial reconstruction. To evaluate prediction accuracy, we focus on one test sensor location equipped with instrumentation (represented by the solid black line) and compare results. The top panel displays the CF strain signal at the selected test sensor. Gray signals correspond to data outside the training and prediction windows. The measured data within these windows are shown as black dashed lines, split by the training and prediction intervals. In the training window, the blue signal represents DeepVIVONet's outputs at the test sensor location, using signals from 3 observer sensors as inputs and data from all instrumented sensors as labels. In the prediction window, the red signal shows DeepVIVONet's outputs at the test sensor location, based solely on signals from the 3 observer sensors. Notably, both the blue and red predictions align well with the true black dashed signals. To gain deeper insights, the middle panel presents the Fast Fourier Transform (FFT) of the signals in the training and prediction windows, highlighting their frequency-domain characteristics. This panel demonstrates that the model not only matches the signal in the time domain but also learns its inherent frequency properties. To further illustrate the model's performance, the red section within the prediction window is magnified. The bottom panel clearly shows how closely the predicted results match the true signal in future time steps, underscoring the model's accuracy during this critical period. From these plots, we observe that our model delivers accurate predictions across a range of CF velocities. Additionally, Figure \ref{fig: DeepVIVONet shear 2430 IL} showcases the prediction of IL strain signals for the "test2430" case, providing deeper insights into the model's predictive performance under varying environmental conditions.

\subsection{Transfer Learning}
\begin{figure}[h!]
\centering
\includegraphics[clip, trim = 0cm 0cm 0cm 0cm,width=1\linewidth]{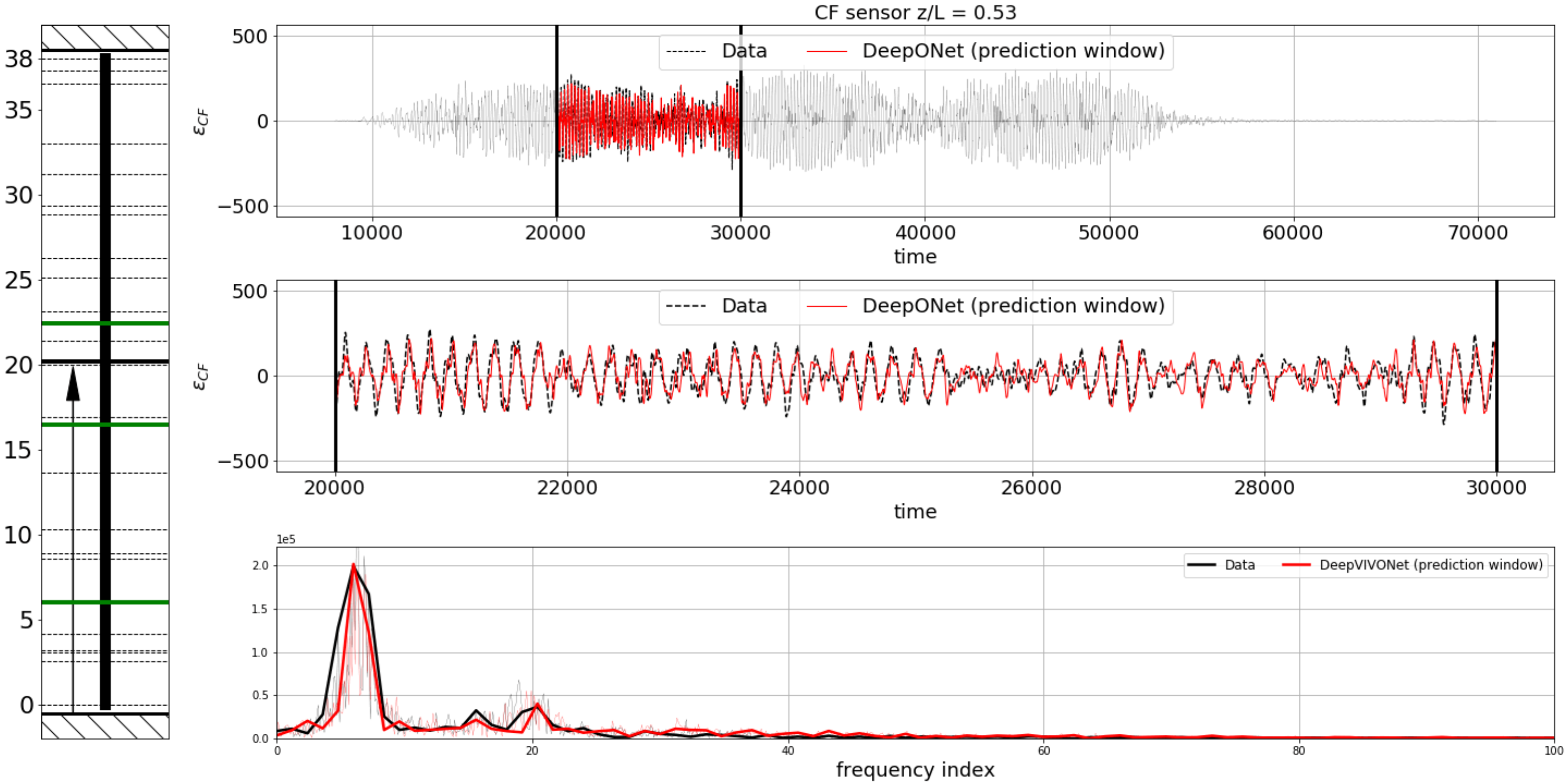}
\caption{DeepVIVONet: Prediction of CF strain for shear flow case ``test2420" using the trained network on case ``test2430". Left panel: the location of all sensors (dashed line), observers (green lines), and test sensor (arrow). Top panel: data (black) and DeepONet prediction (red) in the prediction window. Middle panel: the zoomed-in plot of the prediction window. Bottom panel: the FFT of data and DeepONet.}
\label{fig: DeepVIVONet shear 2420}
\end{figure}

In VIV, the dynamics of marine risers changes as the maximum inflow  velocity changes, and thus it is not trivial for a predictive model to accurately perform predictions across various velocity ranges.  Here, we explore the generalization of our model in extrapolating to different dynamic conditions without the need for full re-training. We train our model by using the CF data of case ``test2430," which has a maximum flow velocity of \( U = 1.50 \, \text{m/s} \). We saved the parameters after this training and then use the trained parameters to predict the dynamics for a different case, ``test2420," characterized by a slightly lower maximum velocity of \( U = 1.40 \, \text{m/s} \). The results are shown in Fig. \ref{fig: DeepVIVONet shear 2420}.

\section{Results on Sensor Placement Optimization}
The architecture of DeepVIVONet is explaind in section \ref{subsec: DeepVIVONet architecture}, where the input to the model is the observer sensors. We have noticed that there exists an optimal observer location, by selecting which the training/testing/forecasting performance improves. Here, we use our DeepVIVONet model as a surrogate for the VIV problem that can obtain the optimal location of observers. First, we explain the setup for sensor placement and then show the optimal locations we obtain. We also compare our results with another method based on proper orthogonal decomposition (POD).

\subsection{Sensor placement}

\begin{figure}[h!]
\centering
\includegraphics[clip, trim = 0cm 0cm 0cm 0cm, width=1\linewidth]{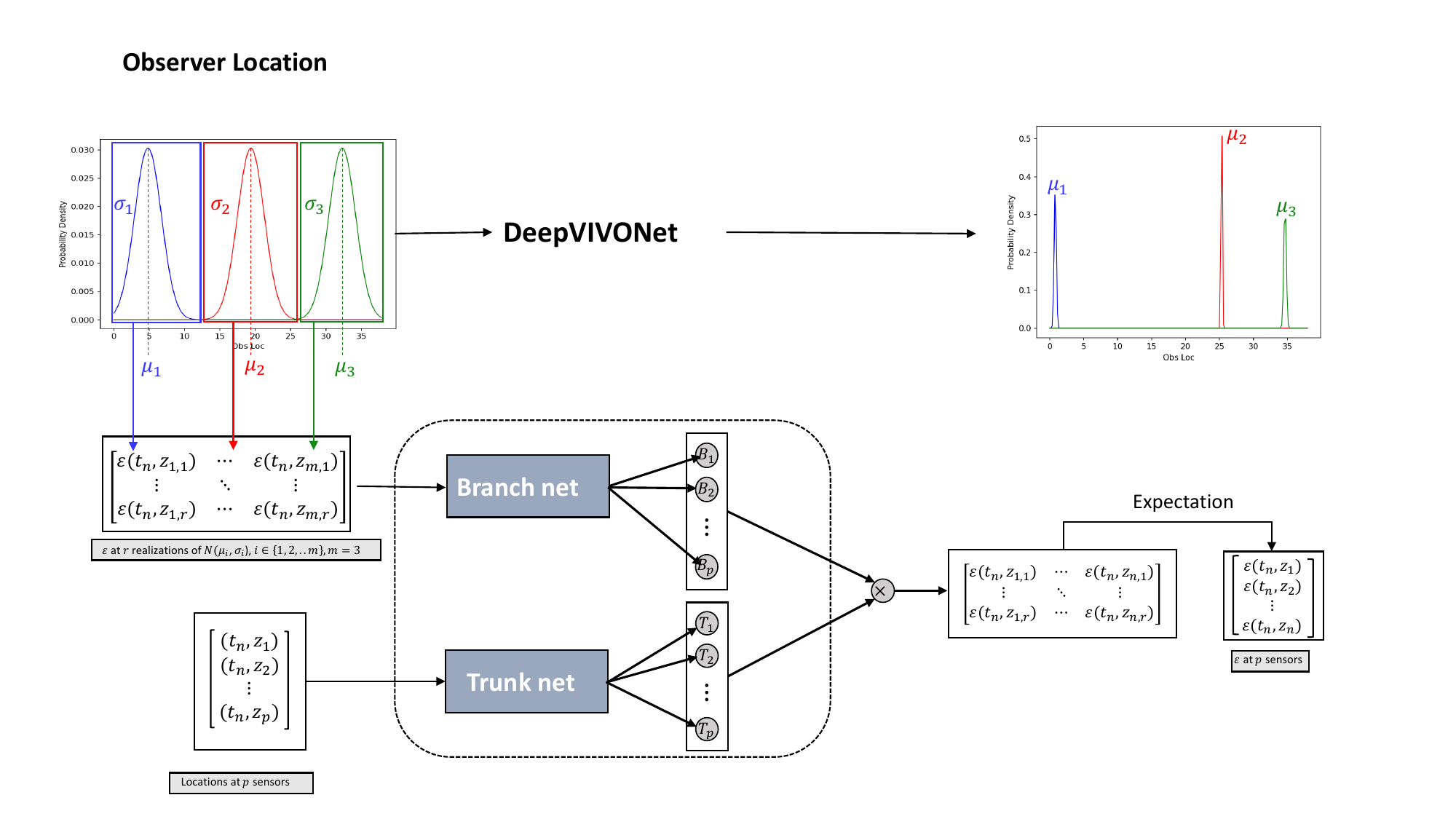}
\caption{Observer location optimization. Here, r realizations are sampled from each observer location distribution, with strains at these points being input to the branch net. Another input to the trunk net is all sensor locations to be reconstructed. The expectation of the output is the strain at all sensor locations with input observer sensors at their optimal value.}
\label{fig: DeepVIVONet-observer}
\end{figure}

To further improve the approximation from sparse signal measurements, DeepVIVONet is also used to select optimal locations for placing the $m$ observer sensors. The locations, denoted as ${z_1, z_2, \cdots, z_m}$, are modeled as learnable distributions characterized by their means ${\mu_1, \mu_2, \ldots, \mu_m}$ and standard deviations ${\sigma_1, \sigma_2, \cdots, \sigma_m}$. These parameters are indicative of the centrality and spread of the sensor locations within the monitored domain. In this case, at each time step, we have $$\mathcal{G_{\theta, \lambda}}(\varepsilon)(t_n, z^*) = \sum_{k=1}^{P} {B_k}(\lbrace \varepsilon(t_n, z_1(\mu_1, \sigma_1)), \cdots, \varepsilon(t_n, z_m(\mu_m, \sigma_m)) \rbrace){T_k}(t_n, z^{\star}) + {B_0}.$$  The optimization process involves dynamically adjusting these parameters to hone in on the most informative sensor placements. The standard deviations ${\sigma_1, \sigma_2, \cdots, \sigma_m}$ are expected to converge towards minimal values as the means ${\mu_1, \mu_2, \cdots, \mu_m}$ approach their optimal locations. This indicates a reduction in uncertainty and an increase in the precision of the sensor array configuration. The training parameters $\lambda$, for location distributions, and $\theta$ for all other parameters to learn the reconstruction mapping, are alternately optimized, ensuring the locations are not overridden by the other reconstruction parameters. Figure \ref{fig: DeepVIVONet-observer} illustrates the progression of these optimizations, depicting how the observer location distributions evolve towards optimal.

The sensor placement optimization is performed using data from the NDP CF case ``test2430". To facilitate spatial analysis and optimize observer locations, the dataset underwent a preprocessing step. This step involved discretizing a reconstructed dataset at 500 equidistant points across the spatial domain to ensure a dense representation in space. Subsequently, for the purposes of model training and testing, data from 100 equidistant locations were selected to form the training set, and data from 50 equidistant points were designated for the test set.

\subsection{Proper Orthogonal Decomposition (POD) Analysis}
To optimize observer locations, Proper Orthogonal Decomposition (POD) was usually employed. POD analyzes the structure of the strain signal $\mathbf{\varepsilon (x, t)}$, decomposing it into deterministic spatial functions $\phi_{k}(x)$ modulated by random time coefficients $a_k(t)$:
$$\varepsilon (x, t) = \sum_{k=1}^{\infty} a_k(t)\phi_k(x).$$
In \cite{yildirim2009efficient} it was shown that the most energetic ocean dynamics could be described by only a few POD modes, and they proposed a method based on extrema of the POD spatial modes for observer placement. Here, by averaging the training samples over the spatial domain and correlating them with each other along the time samples: 
$$E = \begin{pmatrix}
\varepsilon(x_1, t_1) & \cdots & \varepsilon(x_{100}, t_1) \\
\vdots & & \vdots \\
\varepsilon(x_1, t_{250}) & \cdots & \varepsilon(x_{100}, t_{250}) \\
\end{pmatrix},$$
with 100 spatial elements, and 100 time samples, we can obtain the covariance matrix and  extract the eigenvalues $\lambda_1, \cdots, \lambda_{100}$ and corresponding eigenvectors $\phi_1, \cdots, \phi_{100}$. By ordering these based on the descending eigenvalues, we can identify the first five dominant eigenmodes, and we aim to also use the extrema for observer placement.

\begin{figure}[h!]
\centering
\includegraphics[clip, trim = 0cm 0cm 0cm 0cm, width=0.9\linewidth]{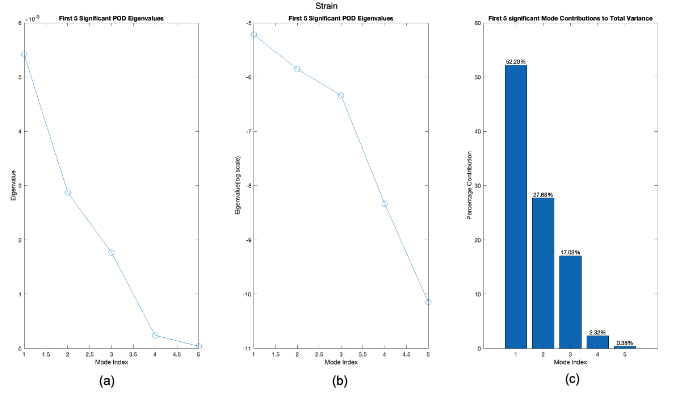}
\caption{Eigenvalues of the first 5 dominant eigenmodes and their percentage contribution to the total energy. (a) Eigenvalues of the first 5 dominant eigenmodes. (b) Eigenvalues of the first 5 dominant eigenmodes plotted in log scale. (c) Percentage contribution of these eigenmodes to the total variance, $\frac{\lambda_i}{\sum_{k=1}^{100} \lambda_k} \cdot 100 \%.$}
\label{fig: POD mode contri}
\end{figure}

\begin{figure}[h!] 
    \centering
    \begin{subfigure}[b]{0.31\textwidth}
        \centering
        \includegraphics[clip, trim = 0cm 0cm 0cm 0cm, width=1\linewidth]{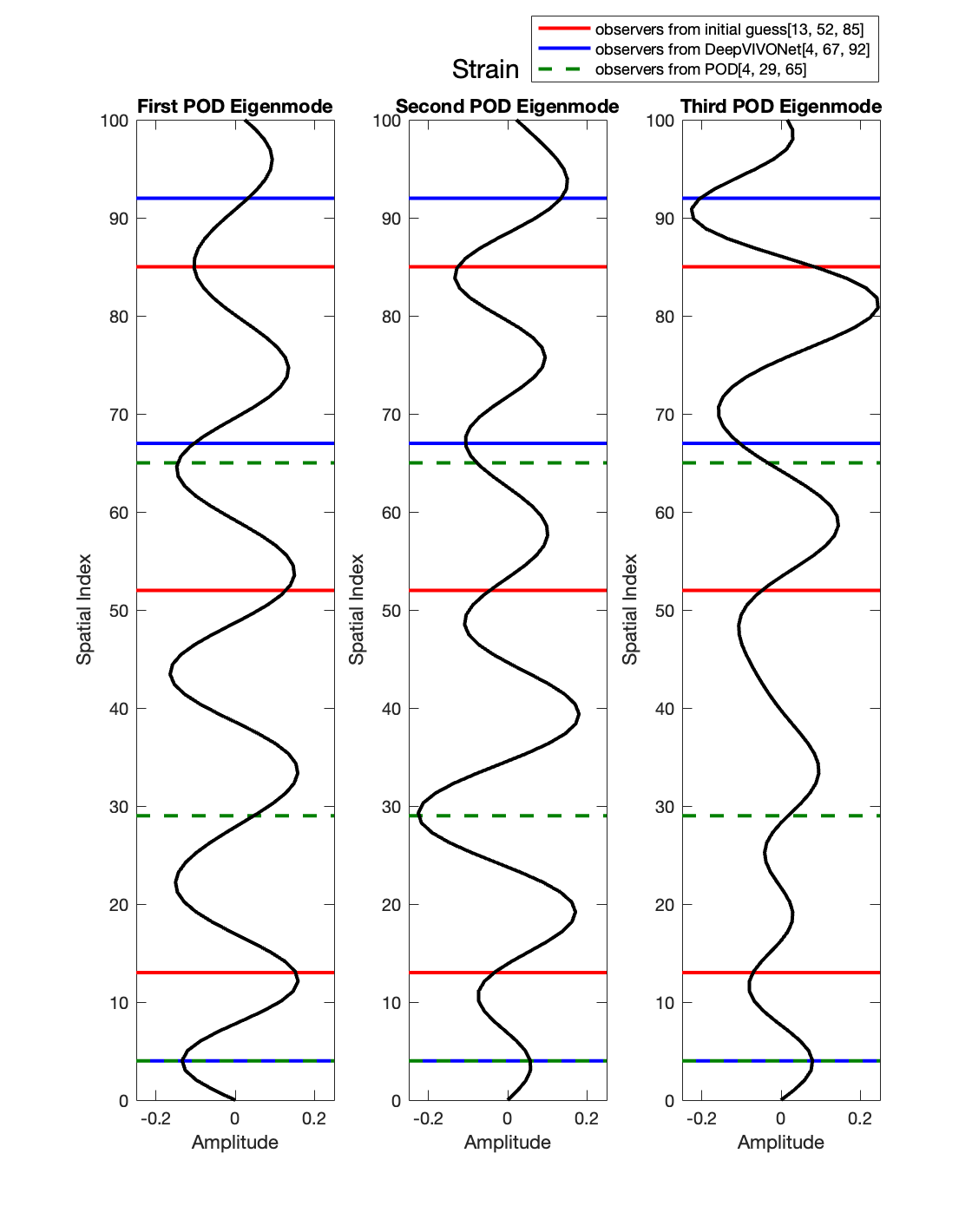}
        \caption{POD chosen locations at 4, 29, 65.}
        \label{fig: POD 4 29 65}
    \end{subfigure}
    \hspace{3pt}
    \begin{subfigure}[b]{0.31\textwidth}
        \centering
        \includegraphics[clip, trim = 0cm 0cm 0cm 0cm, width=1\linewidth]{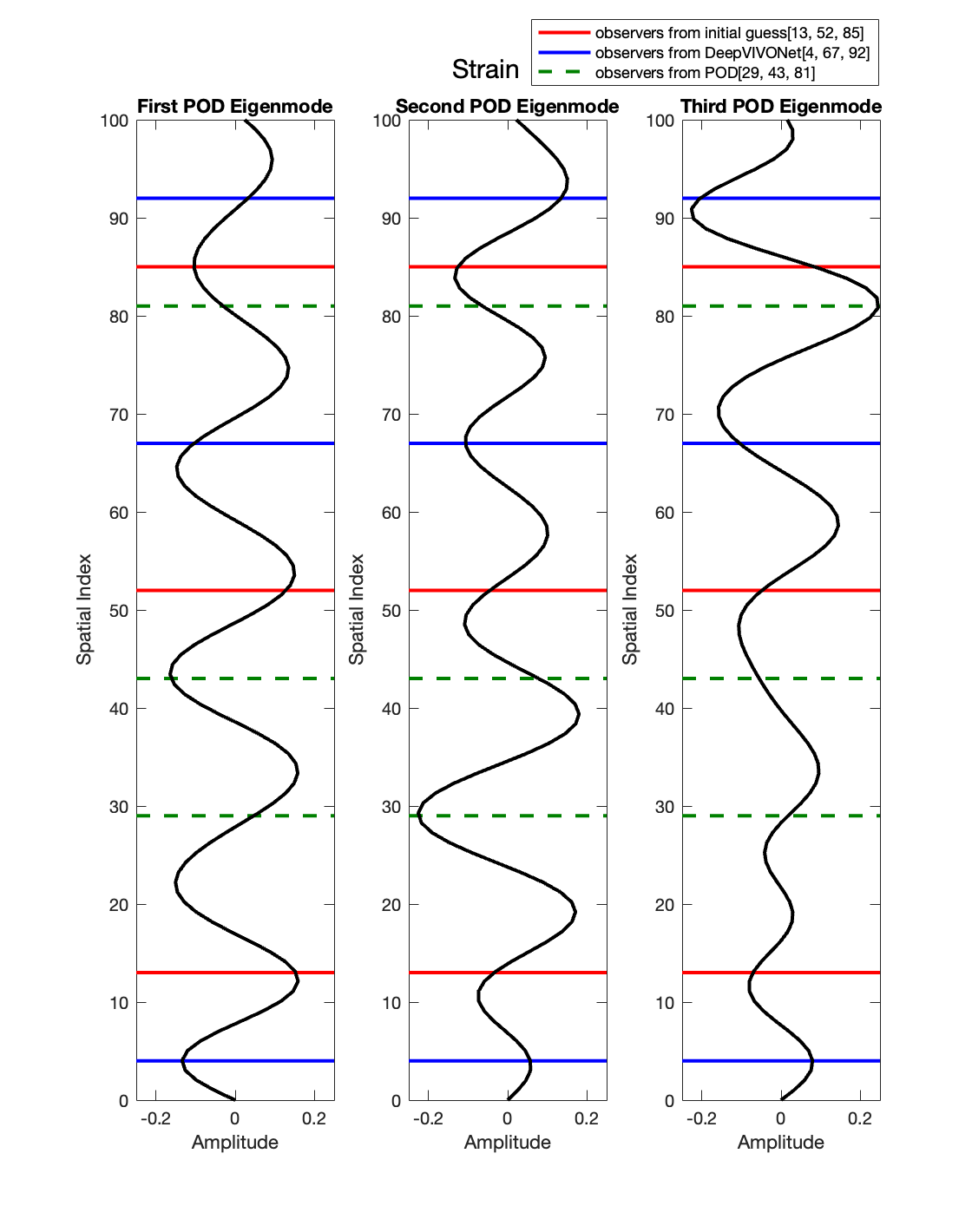}
        \caption{POD chosen locations at 29, 43, 81.}
        \label{fig: POD 29 43 81}
    \end{subfigure}
    \hspace{3pt}
    \begin{subfigure}[b]{0.31\textwidth}
        \centering
        \includegraphics[clip, trim = 0cm 0cm 0cm 0cm, width=1\linewidth]{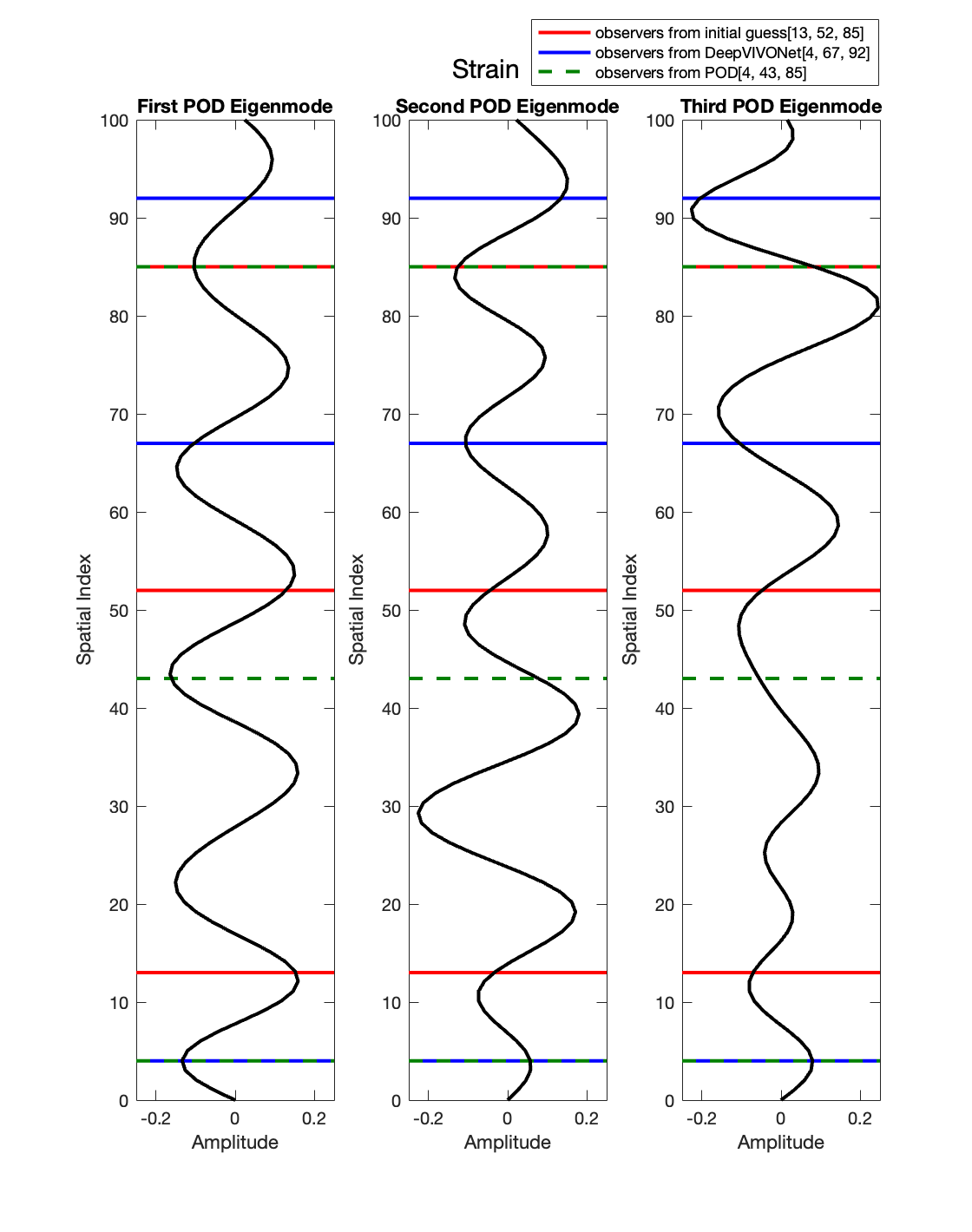}
        \caption{POD chosen locations at 4, 43, 85.}
        \label{fig: POD 4 43 85}
    \end{subfigure}
    
    \caption{First 3 dominant POD modes with observer locations indicated. The locations are random initial guesses from DeepVIVONet, learned final locations from DeepVIVONet and 3 different location combinations chosen manually by observing POD modes.}
    \label{fig: POD modes selection}
\end{figure}

Figure \ref{fig: POD mode contri}c illustrates the percentage contribution of these modes, where the first eigenmode alone accounts for over 50\% of the total variance, and the first three eigenmodes together account for 96.91\% of the variance. Based on these findings, only the locations corresponding to the first three eigenmodes were considered for observer placement. Figure \ref{fig: POD modes selection} displays these dominant modes, characterized by multiple local maxima and minima. Given the complexity of the modes, determining the optimal observer locations only by observing the eigenmodes themselves poses challenges. One way is to select two local extrema from the first mode and one from the second mode, as visualized in Figure \ref{fig: POD 4 29 65}, since the first mode account for more than half of the total variance. Alternative ways included selecting one local extrema from each of the first three modes, as shown in Figure \ref{fig: POD 29 43 81}, and a combination that encompasses locations at or near local extrema or minor local extrema across all three modes, depicted in Figure \ref{fig: POD 4 43 85}.

\subsection{Prediction of Observer Locations}
\begin{figure}[h!]
\centering
\includegraphics[clip, trim = 0cm 0cm 0cm 0cm, width=0.6\linewidth]{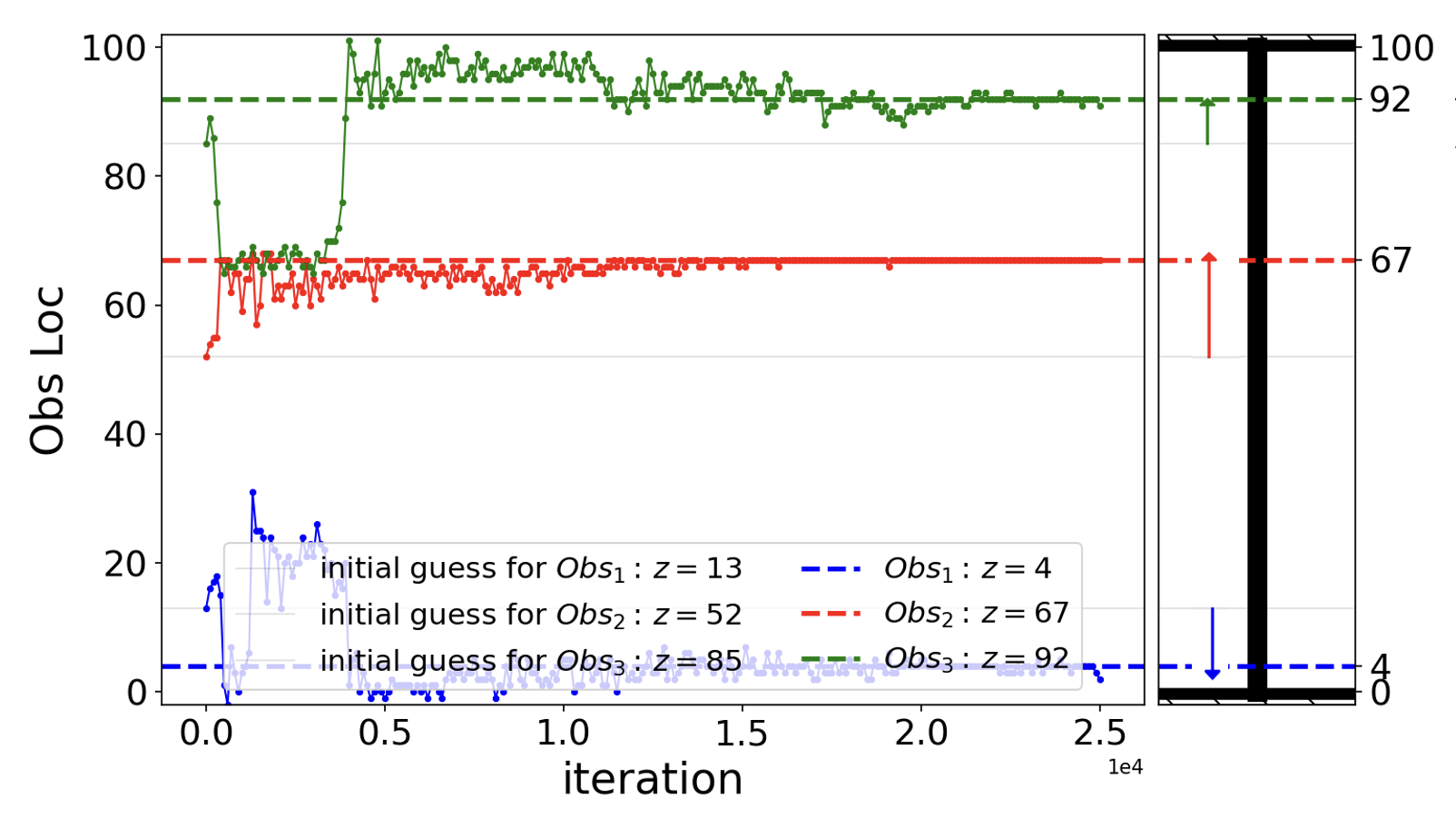}
\caption{DeepVIVONet observer location variation during iterations. Every 100 iterations are saved here as 1 iteration.}
\label{fig: DeepVIVONet observer iter}
\end{figure}

In this section we detail the dynamics of observer location optimization using DeepVIVONet during the training process. The training involved a total of 500,000 iterations, wherein randomly guessed initial observer locations served as starting points. Specifically, the process is started by doing 60 iterations using the initial guesses and 40 iterations adjusting the input locations, and then use the new input locations for the next 60 iterations. The 60-40 alternating process continues during training, aiming to refine and optimize observer placement. Figure \ref{fig: DeepVIVONet observer iter} illustrates an example of how these locations varied significantly at the onset of training but began to converge to specific points as the iterations progressed. This convergence highlights  the optimization algorithm's effectiveness in identifying strategically beneficial sensor placements for accurate strain signal prediction.

\begin{figure}[h!]
\centering
\includegraphics[clip, trim = 0cm 0cm 0cm 0cm, width=1\linewidth]{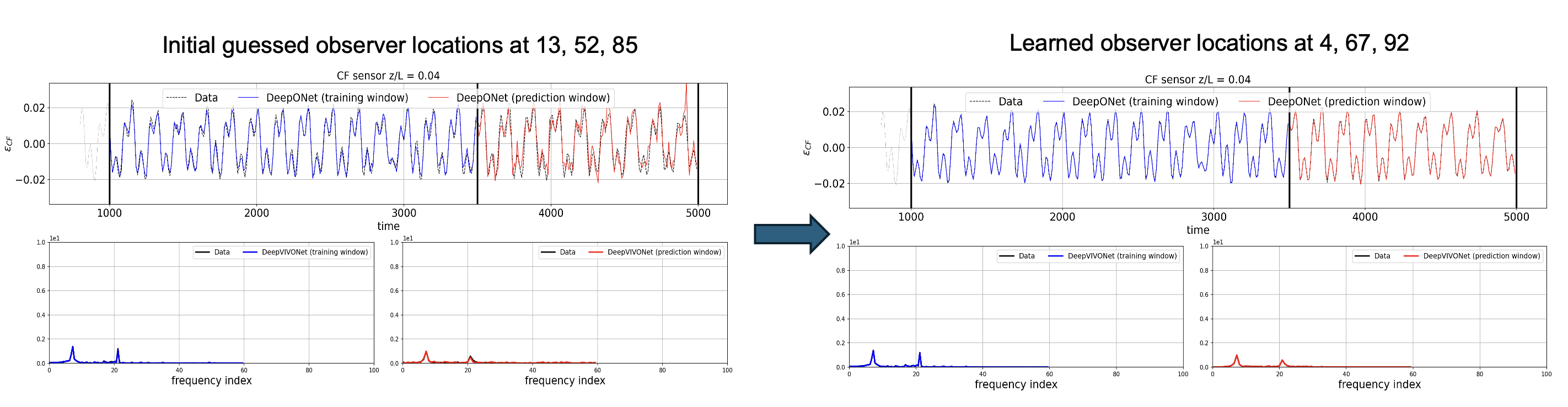}
\caption{DeepVIVONet predictions at a certain reconstructed location using observer locations at initial guesses and at learned observer locations. Left: Prediction using randomly selected initial guesses of 3 observer locations. Right: Prediction using the learned 3 observer locations.}
\label{fig: DeepVIVONet observer loc time}
\end{figure}

\begin{figure}[h!]
\centering
\includegraphics[clip, trim = 0cm 0cm 0cm 0cm, width=0.9\linewidth]{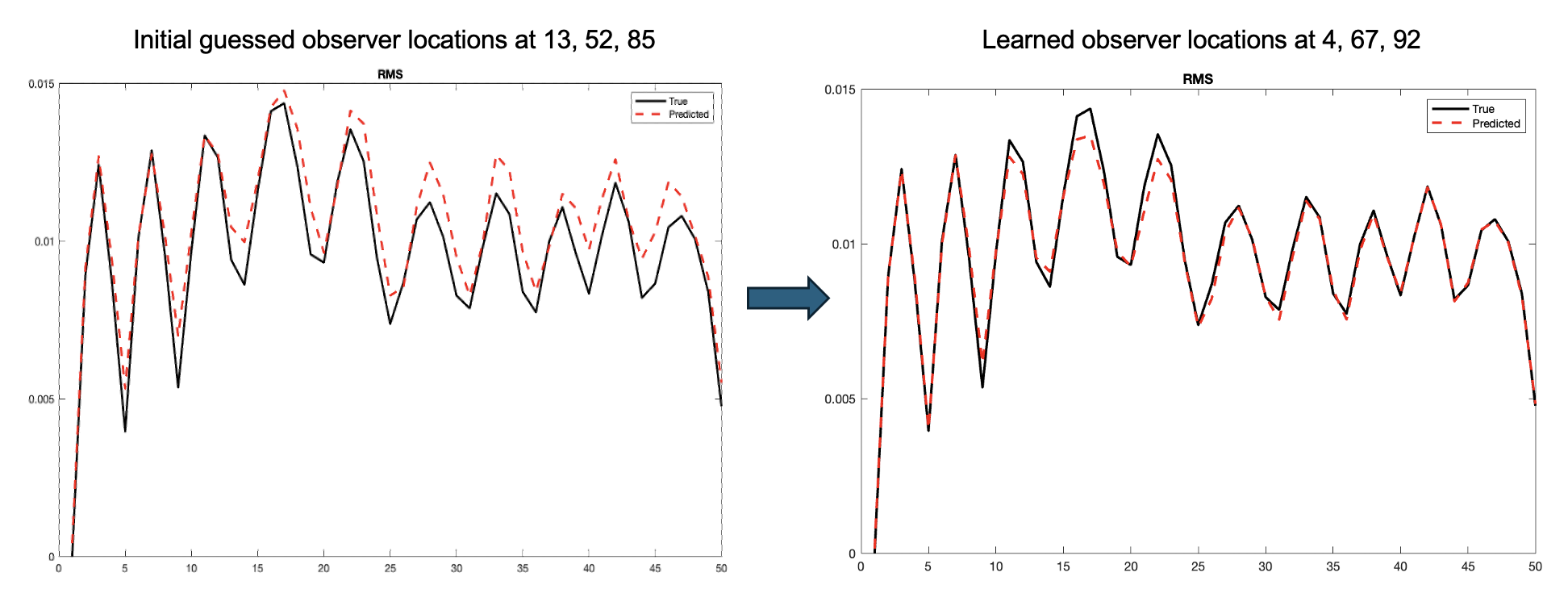}
\caption{Reconstructed RMS for prediction window using data from 3 observers. Left: Prediction using randomly selected initial guesses of 3 observer locations. Right: Prediction using learned 3 observer locations.}
\label{fig: DeepVIVONet observer rms}
\end{figure}

Further insights are provided in Figure \ref{fig: DeepVIVONet observer loc time}, which compares the DeepVIVONet predictions at a particular reconstructed location using both the initial and learned observer locations. The comparison reveals that predictions using the learned locations align more closely with the true strain signals, demonstrating improved accuracy in both the training and prediction windows. Additionally, the Root Mean Square (RMS) strains of all reconstructed locations are depicted in Figure \ref{fig: DeepVIVONet observer rms}. This metric confirms that utilizing data from the learned observer locations results in superior reconstruction quality, underscoring the value of our observer optimization approach in enhancing the model's predictive performance.

\subsection{Comparison of optimization results with POD}
Here, we compare the reconstruction results using the observer location optimization method from DeepVIVONet and POD. Table \ref{obs mse} compares the mean squared error (MSE) of reconstruction in the prediction window for using initial guessed locations, locations learned by DeepVIVONet, and three observer location combinations derived from POD for training. The observer locations optimized by DeepVIVONet consistently delivered superior results compared to other methods. Although one combination from the POD analysis also yielded favorable outcomes, selecting effective combinations from POD can be challenging while significant variations in MSE occur across different POD selections.

\begin{table}[h!]
\centering
\caption{Mean squared error in prediction window using different observer locations for training.}\label{obs mse}
\begin{tabular}{cc}
\hline
Observer locations                                                                     & MSE in prediction window \\ \hline
\begin{tabular}[c]{@{}c@{}}13 52 85\\ (Initial guessed locations)\end{tabular}         & 3.745e-05                \\
\begin{tabular}[c]{@{}c@{}}4 67 92\\ (Learned locations from DeepVIVONet)\end{tabular} & 1.290e-06                \\
\begin{tabular}[c]{@{}c@{}}4 29 65\\ (Locations from POD)\end{tabular}                 & 9.838e-06                \\
\begin{tabular}[c]{@{}c@{}}29 43 81\\ (Locations from POD)\end{tabular}                & 1.294e-06                \\
\begin{tabular}[c]{@{}c@{}}4 43 85\\ (Locations from POD)\end{tabular}                 & 5.410e-06           \\ \hline
\end{tabular}
\end{table}

\section{Conclusion}

In this study, DeepVIVONet, a DeepONet-based architecture, was employed to predict the dynamic behavior of marine risers subjected to vortex-induced vibrations (VIV). The model successfully resolves the complex VIV responses. Our application of transfer learning showcases its effectiveness in adapting to variations in fluid dynamics, particularly in environments with differing maximum velocities. This adaptability enhances the model's utility across a range of operational conditions, making it a valuable tool for offshore engineering applications. Furthermore, the introduction of an optimized approach for the placement of observer sensors marks an advancement in experimental methodologies. By optimizing sensor locations, we have not only increased the efficiency of data collection but also substantially reduced the associated costs. This methodological innovation facilitates more economical and practical deployment of monitoring systems in offshore settings.

DeepVIVONet sets a new benchmark in the field by improving the efficiency of predictions concerning the operational integrity of offshore structures. Future research will aim to expand the model's capabilities by integrating more diverse environmental variables. Such advancements will likely increase the model’s robustness and applicability, contributing broadly to the field of fluid-structure interaction.

\section*{Acknowledgements}
The first and fourth authors acknowledge with gratitude the support by MURI grant (FA9550-20-1-0358). The third author acknowledges with gratitude support from Shell, ExxonMobil, ABS, and Subsea 7.

\appendix
\newpage

\section{Explanation of NDP dataset cases and instrumented sensor locations}
\label{App: NDP}

The riser is a 38 meter pipe with pinned--pinned end conditions \cite{henning2005ndp}. Table \ref{tab: NDP cases details} shows the overview of strain sensor instrumentation and test cases for the NDP dataset. IL and CF stand for in--line and cross flow motions. The sensors No. with acc have accelerometers too. 
\begin{table}[h!]
\centering
\begin{tabular}{
>{\ttfamily\raggedright}p{8cm}
>{\sffamily}p{5cm}
}
\toprule
\multicolumn{1}{c}{\bfseries\sffamily Riser instrumentation locations} & 
\multicolumn{1}{c}{\bfseries\sffamily NDP data case numbers} 
\\
& \\[-2ex]
\vspace{-15pt}
\scalebox{0.7}{
\begin{tabular}[t]{|c c c|}
\toprule
\multicolumn{1}{l}{\bfseries\sffamily No.} 
& \multicolumn{1}{l}{\bfseries\sffamily Instrumentation} 
& \multicolumn{1}{l}{\bfseries\sffamily Dis. from riser top} \\
\midrule
1 & IL & 1.143 \\ 
2 & IL/CF & 2.555 \\ 
3 & IL/CF & 3.084 \\ 
4 & IL/CF &  3.224 \\ 
5 & acc IL/CF &  4.155\\ 
6 & IL &  5.759\\ 
7 & IL/CF & 6.030 \\ 
8 & IL &  7.664\\ 
9 & IL &  8.216\\ 
10 & acc IL/CF & 8.609 \\ 
11 & IL/CF &  8.889\\ 
12 & IL &  9.703\\ 
13 & IL/CF &  10.285\\ 
14 & acc IL/CF &  13.676\\ 
15 & IL & 13.772 \\ 
16 & IL &  15.393\\ 
17 & IL/CF &  16.452\\ 
18 & IL &  16.547\\ 
19 & acc IL/CF & 16.891 \\ 
20 & IL/CF & 19.997 \\ 
21 & IL/CF &  20.193\\ 
22 & acc IL/CF & 21.393 \\ 
23 & IL &  21.603\\ 
24 & IL/CF &  22.460\\ 
25 & IL/CF &  23.165\\ 
26 & acc IL/CF &  25.153\\ 
27 & IL & 25.442 \\ 
28 & IL/CF & 26.254 \\ 
29 & IL & 27.532 \\ 
30 & IL &  28.698\\ 
31 & acc IL/CF & 28.863 \\ 
32 & IL/CF & 29.365 \\ 
33 & IL/CF & 31.191 \\ 
34 & IL & 31.415 \\ 
35 & IL & 32.796 \\ 
36 & acc IL/CF & 33.005 \\ 
37 & IL & 35.549 \\ 
38 & IL/CF & 36.559 \\ 
39 & IL & 36.824 \\ 
40 & IL/CF & 37.332 \\ 
\bottomrule
\end{tabular} 
}
& 
\vspace{-15pt}
\scalebox{0.8}{
\begin{tabular}[t]{|c c c|}
\toprule
\multicolumn{1}{l}{\bfseries\sffamily $U(m/s)$} 
& \multicolumn{1}{l}{\bfseries\sffamily Uniform} 
& \multicolumn{1}{l}{\bfseries\sffamily Shear} \\
\midrule
0.3 & 2010 & 2310 \\ 
0.4 & 2020 & 2320 \\ 
0.5 & 2030 & 2330 \\ 
0.6 & 2040 & 2340 \\ 
0.7 & 2050 & 2350 \\ 
0.8 & 2060 & 2360 \\ 
0.9 & 2070 & 2370 \\ 
1.0 & 2080 & 2380 \\ 
1.1 & 2090 & 2390 \\ 
1.2 & 2100 & 2400 \\ 
1.3 & 2110 & 2410 \\ 
1.4 & 2120 & 2420 \\ 
1.5 & 2130 & 2430 \\ 
1.6 & 2140 & 2440 \\ 
1.7 & 2150 & 2450 \\ 
1.8 & 2160 & 2460 \\ 
1.9 & 2170 & 2470 \\ 
2.0 & 2182 & 2480 \\ 
2.1 & 2191 & 2490 \\ 
2.2 & 2201 & 2500 \\ 
2.3 & 2210 & 2510\\ 
2.4 & 2220 & 2520 \\ 
\bottomrule
\end{tabular} 
}
\\
\bottomrule
\end{tabular}
\caption{NDP data set details: riser instrumentation locations (left) and case numbers (right). }
\label{tab: NDP cases details}
\end{table}

\bibliography{ref}

\begin{thebibliography}{10}
\expandafter\ifx\csname url\endcsname\relax
  \def\url#1{\texttt{#1}}\fi
\expandafter\ifx\csname urlprefix\endcsname\relax\def\urlprefix{URL }\fi
\expandafter\ifx\csname href\endcsname\relax
  \def\href#1#2{#2} \def\path#1{#1}\fi

\bibitem{bourguet2011vortex}
R.~Bourguet, G.~E. Karniadakis, M.~S. Triantafyllou, Vortex-induced vibrations of a long flexible cylinder in shear flow, Journal of Fluid Mechanics 677 (2011) 342--382.

\bibitem{wang2021large}
Z.~C. Wang, D.~X. Fan, M.~S. Triantafyllou, G.~E. Karniadakis, A large-eddy simulation study on the similarity between free vibrations of a flexible cylinder and forced vibrations of a rigid cylinder, Journal of Fluids and Structures (2021) 101.

\bibitem{zheng2022flow}
H.~X. Zheng, J.~S. Wang, Flow-induced vibration of flexible cylinders covered by fixed fairings with different chord-thickness ratios, Marine Structures 86 (2022).

\bibitem{sun2021accurate}
P.~N. Sun, D.~Le~Touze, G.~Oger, A.~M. Zhang, An accurate {FSI-SPH} modeling of challenging fluid-structure interaction problems in two and three dimensions, Ocean Engineering (2021) 221.

\bibitem{lin2019numerical}
K.~Lin, J.~S. Wang, Numerical simulation of vortex-induced vibration of long flexible risers using a {SDVM-FEM} coupled method, Ocean Engineering 172 (2019) 468--486.

\bibitem{jin2023numerical}
G.~Q. Jin, Z.~Zong, Z.~Sun, L.~Zou, H.~Wang, Numerical analysis of vortex-induced vibration on a flexible cantilever riser for deep-sea mining system, Marine Structures 87 (2023).

\bibitem{williamson1996vortex}
C.~H.~K. Williamson, R.~Govardhan, Vortex dynamics in the cylinder wake, Annual Review of Fluid Mechanics 28 (1996) 477--539.

\bibitem{sarpkaya2006viv}
T.~Sarpkaya, A critical review of the intrinsic nature of vortex-induced vibrations, Journal of Fluids and Structures 22~(6-7) (2006) 703--730.

\bibitem{bearman1984vortex}
P.~W. Bearman, Vortex shedding from oscillating bluff bodies, Annual Review of Fluid Mechanics 16 (1984) 195--222.

\bibitem{wang2020review}
J.~S. Wang, D.~X. Fan, K.~Lin, A review on flow-induced vibration of offshore circular cylinders, Journal of Hydrodynamics 32 (2020) 415--440.

\bibitem{ma2022flexible}
L.~X. Ma, K.~Lin, D.~X. Fan, J.~S. Wang, M.~S. Triantafyllou, Flexible cylinder flow-induced vibration, Physics of Fluids 34 (2022).

\bibitem{jin2020sympnets}
P.~Jin, Z.~Zhang, A.~Zhu, Y.~Tang, G.~E. Karniadakis, Sympnets: Intrinsic structure-preserving symplectic networks for identifying hamiltonian systems, Neural Networks 132 (2020) 166--179.

\bibitem{zhang2022aoslo}
Q.~Zhang, K.~Sampani, M.~Xu, S.~Cai, Y.~Deng, H.~Li, J.~K. Sun, G.~E. Karniadakis, Aoslo-net: A deep learning-based method for automatic segmentation of retinal microaneurysms from adaptive optics scanning laser ophthalmoscopy images, Translational Vision Science \& Technology 11~(8) (2022) 7--7.

\bibitem{toscano2023teeth}
J.~D. Toscano, C.~Zuniga-Navarrete, W.~D. Jo~Siu, L.~J. Segura, H.~Sun, Teeth mold point cloud completion via data augmentation and hybrid rl-gan, Journal of Computing and Information Science in Engineering 23~(4) (2023) 041008.

\bibitem{daneker2022systems}
M.~Daneker, Z.~Zhang, G.~E. Karniadakis, L.~Lu, Systems biology: Identifiability analysis and parameter identification via systems-biology informed neural networks, arXiv preprint arXiv:2202.01723 (2022).

\bibitem{raissi2019physics}
M.~Raissi, P.~Perdikaris, G.~E. Karniadakis, Physics-informed neural networks: A deep learning framework for solving forward and inverse problems involving nonlinear partial differential equations, Journal of Computational Physics 378 (2019) 686--707.

\bibitem{zhang2021integrated}
S.~Zhang, J.~Ponce, Z.~Zhang, G.~Lin, G.~Karniadakis, An integrated framework for building trustworthy data-driven epidemiological models: Application to the covid-19 outbreak in new york city, PLOS Computational Biology 17~(9) (2021) 1--29.

\bibitem{ovadia2023realtime}
O.~Ovadia, V.~Oommen, A.~Kahana, A.~Peyvan, E.~Turkel, G.~E. Karniadakis, Real-time inference and extrapolation via a diffusion-inspired temporal transformer operator ({DiTTO}), arXiv preprint arXiv:2307.09072 (2023).

\bibitem{zhang2022analyses}
E.~Zhang, M.~Dao, G.~E. Karniadakis, S.~Suresh, Analyses of internal structures and defects in materials using physics-informed neural networks, Science Advances 8~(7) (2022) eabk0644.

\bibitem{chen2020physics}
Y.~Chen, L.~Lu, G.~E. Karniadakis, L.~Dal~Negro, Physics-informed neural networks for inverse problems in nano-optics and metamaterials, Optics Express 28~(8) (2020) 11618--11633.

\bibitem{shukla2024deep}
K.~Shukla, V.~Oommen, A.~Peyvan, M.~Penwarden, N.~Plewacki, L.~Bravo, A.~Ghoshal, R.~M. Kirby, G.~E. Karniadakis, Deep neural operators as accurate surrogates for shape optimization, Engineering Applications of Artificial Intelligence 129 (2024) 107615.

\bibitem{oommen2023integrating}
V.~Oommen, A.~Bora, Z.~Zhang, G.~E. Karniadakis, Integrating neural operators with diffusion models improves spectral representation in turbulence modeling, arXiv preprint arXiv:2409.08477 (2024).

\bibitem{peyvan2024riemannonets}
A.~Peyvan, V.~Oommen, A.~D. Jagtap, G.~E. Karniadakis, {RiemannONets}: Interpretable neural operators for {Riemann} problems, Computer Methods in Applied Mechanics and Engineering 426 (2024) 116996.

\bibitem{kharazmi2019variational}
E.~Kharazmi, Z.~Zhang, G.~E. Karniadakis, Variational physics-informed neural networks for solving partial differential equations, arXiv preprint arXiv:1912.00873 (2019).

\bibitem{kharazmi2021hp}
E.~Kharazmi, Z.~Zhang, G.~E. Karniadakis, hp-vpinns: Variational physics-informed neural networks with domain decomposition, Computer Methods in Applied Mechanics and Engineering 374 (2021) 113547.

\bibitem{jagtap2020conservative}
A.~D. Jagtap, E.~Kharazmi, G.~E. Karniadakis, Conservative physics-informed neural networks on discrete domains for conservation laws: Applications to forward and inverse problems, Computer Methods in Applied Mechanics and Engineering 365 (2020) 113028.

\bibitem{cao2023deep}
Q.~Cao, S.~Goswami, T.~Tripura, S.~Chakraborty, G.~E. Karniadakis, Deep neural operators can predict the real-time response of floating offshore structures under irregular waves, Computers \& Structures 290 (2023) 107228.

\bibitem{kharazmi2021inferring}
E.~Kharazmi, D.~Fan, Z.~Wang, M.~S. Triantafyllou, Inferring vortex induced vibrations of flexible cylinders using physics-informed neural networks, Journal of Fluids and Structures 107 (2021) 103367.

\bibitem{kharazmi2021data}
E.~Kharazmi, Z.~Wang, D.~Fan, S.~Rudy, T.~Sapsis, M.~S. Triantafyllou, G.~E. Karniadakis, From data to assessment models, demonstrated through a digital twin of marine risers, in: Offshore Technology Conference, OTC, 2021, p. D031S035R003.

\bibitem{mentzelopoulos2024variational}
A.~P. Mentzelopoulos, D.~Fan, T.~P. Sapsis, M.~S. Triantafyllou, Variational autoencoders and transformers for multivariate time-series generative modeling and forecasting: Applications to vortex-induced vibrations, Ocean Engineering 310, Part 2 (2024) 118639.

\bibitem{lu2021deeponet}
L.~Lu, P.~Jin, G.~Pang, et~al., Learning nonlinear operators via {DeepONet} based on the universal approximation theorem of operators, Nature Machine Intelligence 3 (2021) 218--229.

\bibitem{Hornik1989}
K.~Hornik, M.~Stinchcombe, H.~White, Multilayer feedforward networks are universal approximators, Neural Networks 2~(5) (1989) 359--366.

\bibitem{chen1995universal}
T.~Chen, H.~Chen, Universal approximation to nonlinear operators by neural networks with arbitrary activation functions and its application to dynamical systems, IEEE Transactions on Neural Networks 6~(4) (1995) 911--917.

\bibitem{henning2005ndp}
H.~Braaten, H.~Lie, {NDP} riser high mode {VIV} tests main report, Tech. rep., Norwegian Marine Technology Research Institute (2005).

\bibitem{trygve2005ndpreponse}
T.~Kristansen, H.~Lie, {NDP} riser high mode {VIV} response analysis, Tech. rep., Norwegian Marine Technology Research Institute (2005).

\bibitem{trygve2005ndp}
T.~Kristansen, H.~Lie, {NDP} riser high mode {VIV} tests-modal analysis, Tech. rep., Norwegian Marine Technology Research Institute (2005).

\bibitem{gro2005ndp}
G.~S. Baarholm, {NDP} riser high mode {VIV} tests / fatigue analysis, Tech. rep., Norwegian Marine Technology Research Institute (2005).

\bibitem{yildirim2009efficient}
B.~Yildirim, C.~Chryssostomidis, G.~Karniadakis, Efficient sensor placement for ocean measurements using low-dimensional concepts, Ocean Modelling 27~(3-4) (2009) 160--173.

\end{thebibliography}

\bibliographystyle{elsarticle-num}

\end{document}